\documentclass[10pt]{article}

\usepackage{rotating}  
\usepackage{theorem}   
\usepackage{bm}  
\usepackage{amsmath,amsfonts,amssymb,mathabx}
\usepackage{booktabs}  
\usepackage{pdfsync}   
\usepackage{pdfpages}
\usepackage{float}
\usepackage{threeparttable}
\usepackage{makecell}
\usepackage{adjustbox}
\usepackage{listings}
\usepackage{algorithm}

\usepackage[explicit]{titlesec}
\usepackage{hyphenat}
\usepackage{ragged2e}
\RaggedRight

\usepackage{algorithmic}
\usepackage{multicol}

\usepackage{rotating}  
\usepackage{theorem}   
\usepackage{bm}  
\usepackage{amsmath,amsfonts,amssymb}
\usepackage{booktabs}  
\usepackage{pdfsync}   
\usepackage{graphicx}
\usepackage{latexsym}
\usepackage{amsmath}
\usepackage{amssymb}
\usepackage{fancyvrb}
\usepackage{subfigure}
\usepackage{colortbl}
\usepackage{enumerate}
\usepackage{pifont}
\usepackage{stmaryrd}
\usepackage{textcomp}
\usepackage{fncylab}
\usepackage{multirow}
\usepackage{paralist}
\usepackage{wrapfig}
\usepackage{longtable}
\usepackage{lscape}
\usepackage{hyperref}
\usepackage{lipsum}

\usepackage{garamondx}
\usepackage[T1]{fontenc}
\usepackage{amsmath}
\usepackage{graphicx}
\usepackage{xcolor}

\usepackage{geometry}
\usepackage{fancyhdr}
\usepackage[labelfont={footnotesize,bf} , textfont=footnotesize]{caption}
\makeatletter
\renewcommand\tagform@[1]{\maketag@@@ {\ignorespaces {\footnotesize{\textbf{Equation}}} #1.\unskip \@@italiccorr }}
\makeatother
\titleformat{\section}
  {\normalfont}{\thesection}{1em}{\MakeUppercase{\textbf{#1}}}
\titlespacing\section{0pt}{0pt}{-10pt}
\titleformat{\subsection}
  {\normalfont}{\thesubsection}{1em}{\textit{#1}}
\titlespacing\subsection{0pt}{0pt}{-8pt}

\makeatletter
\newcommand\sixteen{\@setfontsize\sixteen{17pt}{6}}
\renewcommand{\maketitle}{\bgroup\setlength{\parindent}{0pt}
\begin{flushleft}
\sixteen\bfseries \@title
\medskip
\end{flushleft}
\textit{\@author}
\egroup}
\makeatother

\usepackage[sort&compress]{natbib}
\makeatletter
\renewcommand\@biblabel[1]{\textbf{#1.}\hfill}
\makeatother

\bibpunct{}{}{,~}{s}{,}{,}
\colorlet{punct}{red!60!black}
\definecolor{background}{HTML}{EEEEEE}
\definecolor{delim}{RGB}{20,105,176}
\colorlet{numb}{magenta!60!black}

\lstdefinelanguage{json}{
    basicstyle=\normalfont\ttfamily,
    numbers=left,
    numberstyle=\scriptsize,
    stepnumber=1,
    numbersep=8pt,
    showstringspaces=false,
    breaklines=true,
    frame=lines,
    backgroundcolor=\color{background},
    literate=
     *{0}{{{\color{numb}0}}}{1}
      {1}{{{\color{numb}1}}}{1}
      {2}{{{\color{numb}2}}}{1}
      {3}{{{\color{numb}3}}}{1}
      {4}{{{\color{numb}4}}}{1}
      {5}{{{\color{numb}5}}}{1}
      {6}{{{\color{numb}6}}}{1}
      {7}{{{\color{numb}7}}}{1}
      {8}{{{\color{numb}8}}}{1}
      {9}{{{\color{numb}9}}}{1}
      {:}{{{\color{punct}{:}}}}{1}
      {,}{{{\color{punct}{,}}}}{1}
      {\{}{{{\color{delim}{\{}}}}{1}
      {\}}{{{\color{delim}{\}}}}}{1}
      {[}{{{\color{delim}{[}}}}{1}
      {]}{{{\color{delim}{]}}}}{1},
}

\title{Towards Explainable Artificial Intelligence in Banking and Financial Services}

\author{
Ambreen Hanif*$^{a}$ \\ \medskip
$^{a}$Macquarie University, Sydney, Australia \\  \medskip
ambreen.hanif@hdr.mq.edu.au
}

\begin{document}

\vspace*{.01 in}
\maketitle
\vspace{.12 in}

\section*{abstract}

Artificial intelligence (AI) enables machines to learn from human experience, adjust to new inputs, and perform human-like tasks. 
AI is progressing rapidly and is transforming the way businesses operate, from process automation to cognitive augmentation of tasks and intelligent process/data analytics.
However, the main challenge for human users would be to understand and appropriately trust the result of AI algorithms and methods.
In this paper, to address this challenge, we study and analyze the recent work done in Explainable Artificial Intelligence (XAI) methods and tools. We introduce a novel XAI process, which facilitates producing explainable models while maintaining a high level of learning performance.
We present an interactive evidence-based approach to assist human users in comprehending and trusting the results and output created by AI-enabled algorithms.
We adopt a typical scenario in the Banking domain for analyzing customer transactions. We develop a digital dashboard to facilitate interacting with the algorithm results and discuss how the proposed XAI method can significantly improve the confidence of data scientists in understanding the result of AI-enabled algorithms.

\section*{keywords}
Business Process Analytics; Explainable AI; Machine learning

\vspace{.12 in}


\section{introduction}

This section present an overview of the completed study and explains the author's rationale to conduct this research. We will go through the problem we are addressing and discuss the impacts of the proposed approach for the business organizations. 
Furthermore, we present the project's contribution to research and academia to meet the goals and its response to the demands of the finance industry.

\subsection{Overview}
\label{paper_overview}

In the last decade, the world has envisioned tremendous
growth in technology with the improved accessibility of
data, cloud resources, and machine learning (ML) algorithms
evolution. The intelligent system has achieved significant performance
with this growth. The super performance of these
algorithms in various domains has increased the popularity of
artificial intelligence (AI). However, alongside these achievements,
the non-transparent, ambiguity, and inability to expound and
interpret the majority of the state-of-the-art techniques are
considered ethical issues. These flaws in AI algorithms impede
the acceptance of complex ML models in a variety of fields such
as medical, banking and finance, security, and education and
have prompted many concerns about the security and safety of
ML system users. According to the current regulations and policies, these systems must be transparent to meet the right to explanation. Due to a lack of trust in existing ML-based systems, explainable artificial intelligence (XAI)-based methods are gaining popularity. Although neither the domain nor the methods are novel, they are gaining popularity due to their ability to unbox the black box.

This paper highlights the importance of Explainable Artificial Intelligence (XAI) with Artificial Intelligence (AI) development in the financial sector. The financial sector is the primary driver of the country's economy. The excellent performance of the financial sector is an indication of the country's economic growth. One of the key focuses of the financial industry is to increase the return over investment (ROI) within the defined risk appetite and threshold of risk tolerance. In the financial sector, the organisation's risk appetite can be a leading indicator of industry growth. The higher risk appetite and high-risk tolerance indicate that the industry is willing to take the high risk and is tolerant enough to bear the loss beyond the defined risk levels. Despite the higher risk appetite, this sector is lagging in AI adoption. The significant reasons for lagging in AI are the lack of data infrastructure, low ROI in the beginning, and scrutinised cost by the higher management 

A study on Machine learning explainability in finance \cite{Bracke2019MachineAnalysis} has provided insights into the finance industry future and trends. One of the survey's findings is that by 2025 many institutes need to incorporate intelligent algorithms and fact-based information in the system to fulfil the customers' demands.
The banks must adopt the AI and raise the confidence of stakeholders in the financial organisation to analyse as a case study to understand the decisions better to gain insights into the risk associated with the customer.

We investigate the contemporary influences of technology, process analytics~\cite{ProcessAnalyticsB,ProcessAtlas}, and business process on banking. Furthermore, we establish an intelligent evidence-oriented pipeline to extract essential details from a bank transaction classifier and proceed with the data to interactively visualise it in the tuneable dashboard and extract the meaningful feature through the reverse feature engineering component. It would support the AI system of the financial institution to understand the contribution of features and find relevant pieces of evidence through the dashboard interactively. To verify that the system is effective and efficient, the success of the recommended technique is measured.

\subsection{Motivation}

In the financial sector, lending is one of the vast parts that straightforwardly sway the economy. 
According to the Australian Bureau of statistics reports of August 2021~\cite{2021LendingStatistics}, many Australians are worth billions of dollars in credits. Any innovation that can assist a banking sector with a bit of improvement in the organisation's profit from the advances they have or work on in their portions on the lookout would merit a lot of cash. That is why monetary institutions, including banks and private associations, are seeking for ways to improve, develop, and consider incorporating AI into decision-making processes. This way to apply AI is just the initial step and has to go a long way to achieve the goals.

Lending money, a significant source of income in financial institutes, has become an information issue due to the growing user data, making it an appropriate business application for ML. Part of the worth of the loan is attached to the financial soundness of the individual or company. The more information you have about the borrower (and how comparative people have repaid obligations before), the better you can survey their financial soundness. 

The worth of an advance is along these lines attached to evaluations of the value of the insurance (vehicle, home, business, craftsmanship, and so forth), the probable degree of future expansion, and forecasts about by extensive monetary development. AI guarantees that hypothetically it can break down these information sources. It can support the infrastructure to make a sound choice. With the help of AI algorithms, the organisation can calculate the borrower's risk in the risk matrix \cite{garvey1998risk} and automate the scorecard calculations. 

The Inscrutable nature of AI is a hindrance in impedance of AI in the domain of finance. 
Current reservations of investors, and the advancements and success of AI has motivated us to conduct scientific investigations in the existing application of XAI in finance that can support the growth of AI in the industry and convince the investors to invest in intelligent decision systems. 

\subsubsection{Problem Statement}

Artificial intelligence (AI) enables machines to learn from human experience, adjust to new inputs, and perform human-like tasks. 
AI is progressing rapidly and is transforming the way businesses operate, from process automation to cognitive augmentation of tasks and intelligent process/data analytics~\cite{frankCognitive,CognigRS,CognigMBar}.
However, the main challenge for human users would be to understand and appropriately trust the result of AI algorithms and methods.
Focusing on a motivating scenario in the Banking domain, 
to date, the inability of the ML system to explain the generated result is a barrier to the acceptance of these approaches in commercial systems. ML systems are black boxes in nature, and they cannot answer what, how and why to the user. This inability shakes stakeholders' and investors' trust and faces cost scrutinizing to improve the system and increase the risk. 
On the other hand, to improve the financial systems with the growing data, we need the support of intelligent systems to support the decision-making system as the financial system need the help of compliance officers to understand banking data. Still, with the growing data domain, experts cannot cover all the aspects of the data. 

The domain experts have their limitations due to data growth. In contrast, AI system's inability to explain their decisions are two ends that need to meet to fulfil the needs of financial organizations. 
The need of the hour is to have a system in place that can bridge this gap by providing the solution to the limitation of AI-based systems and supporting the domain expert to understand the working of AI system to trust the decisions and proceed with this system. 
This lacking of confidence is the problem at hand we are addressing in this study. 

\subsubsection{Contributions}

In this paper, we study and analyze the recent work done in Explainable Artificial Intelligence (XAI) methods and tools. We introduce a novel XAI process, which facilitates producing explainable models while maintaining a high level of learning performance.
We present an interactive evidence-based pipeline composed of a set
of processes. These steps range from customer data collected from various sources to feature vector and Machine Learning (ML) model prediction generation with interactive visualization. The interactive component helps understand the generated prediction. The generated predictions are helpful to reverse feature engineer the feature vector for the algorithm. The proposed pipeline aims to provide the insights to raise the confidence of business stakeholders in the ML algorithm results and help the data scientist understand the feature contribution for the algorithm decisions, understanding the feature importance. 
The system automates reverse feature engineering, i.e., algorithm decisions and data projections, to extract the importance of features from the predictions. In this study, we take on a scenario for
classifying and analyzing customer bank transaction records to illustrate the presented approach and present the significance of this pipeline to improve the transaction classification engine with the support of reverse feature engineering.

\subsection{Summary and Outline}

This section described the problem that this research focuses on and intends to address. The motivations that lead us to discover this research challenge in the financial industry, as well as our contribution to this paper, have been discussed. The following sections of this paper include the following information:
section \ref{chap:literature} discusses an overview of Explainable Artificial Intelligence (XAI), its evolution and limitations. Application of XAI in time series data as the transaction data is time-series data. Fundamental approaches and the needs of financial organizations are to bring readers attention to the impact of Intelligent algorithms with explainability components on everyday life.

section \ref{chap:Methodology} provides a detailed description of the methodology proposed to achieve the aims and objective of the paper. 

section \ref{chap:motivating} presents elaboration on the problem statement with the motivating scenario. This section also highlights the needs of the proposed approach and 
briefs the reader about the raw dataset details and the experimental setup of the system,
and concludes the methodology with the proposed approach results.

In section \ref{chap:discussion}, We discuss the next steps and propose the future work of the paper.


\section{Background and State-of-the-Art}
\label{chap:literature}

We have divided this section into three sections. The first section discusses the available artificial intelligence based approaches in the literature, in the second section we highlight the explanation techniques used for the time series data. The last section coves the existing interactive tools for the insights into the systems.

\subsection{Explainable Artificial Intelligence (XAI) Methods}

Explainability has evolved as an active research area, with several existing approaches, and due to the non-transparent and inscrutable behaviour of the machine learning algorithms. We generally have little understanding of why AI-based systems make the decisions or exhibit certain behaviours. This ubiquitous nature and inscrutability grow when the system becomes more complex, like deep neural networks complex. So it is inherently difficult to predict which input will contribute to the output and to what extent. This surged the need for explanations to increase the trustability of the algorithms.
We will provide, in this section, a summary of the selected XAI techniques and a list of their strengths, drawbacks, and obstacles. This research is mainly based on the findings of three extensive studies and surveys \cite{Molnar2021InterpretableLearning} \cite{Islam2021ExplainableSurvey} and \cite{Doshi-Velez2017TowardsLearning}. For classification, we are primarily using the survey by Xiao Li. et al. \cite{Li2020AAI}. In their research, they identified two major categories for XAI-based approaches: data-driven approaches and knowledge-driven approaches.
Various post-hoc strategies explain black-box models; one element is conveying the input variables to humans to interpret the systems.
Based on the findings of several investigations, explainability approaches are classified into several categories. These studies
\cite{Doshi-Velez2017TowardsLearning}, \cite{Hall2019ExplainAI}, \cite{Molnar2021InterpretableLearning,attetionRahman}, \cite{Islam2021ExplainableSurvey} referred to three types of explainable systems: intrinsically interpretable techniques, model agnostic techniques, and example-based methods.
We will use the categories listed above and incorporate them into the respective data-driven and knowledge-driven strategies. Studies
\cite{Molnar2021InterpretableLearning} \cite{Andrews1995SurveyNetworks} and \cite{Hall2019ExplainAI} also discuss neural network interpretability and Deep neural network explainability, respectively.
\subsection{Intrinsically Interpretable Methods}
The first and most crucial step towards XAI is to employ models that are straightforward and self-explanatory. In this section, we will look at a few intrinsically interpretable models.

\textbf{Linear Regression}.
These models are thought to be generally interpretable, particularly when their regression coefficients have a clear meaning. The predictions of this technique are a weighted sum of the feature set, as indicated  in section 4.1 by Molnar\cite{Molnar2021InterpretableLearning} . The weighted sum gives the system transparency. The model benefits from the linear nature of the relationship between variables. This characteristic aided these systems in standing out and will be widely accepted by numerous domains including, but not limited to, medical, sociology, sciences, finance, and so on.

\textbf{Logistic Regression}.
Logistic regression is an extension of linear regression that provides a solution to the classification problem. Instead of a straight-line relationship, it squeezes the output between [0,1] to anticipate  the likelihood of  class as an output. The weight indicates the interpretation of the inclination direction. It can be both negative and positive.

\textbf{Decision Tree-Based Models}.
Tree-based models operate on the premise of repeatedly separating data based on specified cutoff values in the system feature set. This property makes them suitable models for predicting interactive features in data where linear and logistic regressions fail.
With trees of manageable height and width, these models will be interpretable. Tree-based models are much more challenging to comprehend for dense trees, and even minor input variations can significantly impact the output because these models are not smooth. Another disadvantage is that they are not as adaptable.
Because they resemble the workings of the sophisticated one, these models are frequently referred to as reverse engineering models. This is also referred to as the 'surrogate model' after the pioneering work~\cite{Craven1995ExtractingNetworks}.

\textbf{Decision Rules}.
Decision rule models are also seen to be simple to understand. Simple if-else statements serve as the system's decision rules, and they can be turned to trees using the terminal, which is a clear win in terms of clarity and explanations. These criteria follow a generic framework to forecast the result and thus require a non-zero positive feature value, with no maximum limit on the criteria.

\textbf{Generalized Linear Models (GLM)}.
The expansion to the linear model is known as the Generalized Linear Model (GLM). The usage of GLM benefits are twofold; one, they preserve the feature's weighted sum and second, they allow for distributions other than Gaussian. The latter was not possible with the linear model. Finally, GLM relates the predicted mean of worked-out distribution with the weighted-sum via potentially nonlinear function.

\textbf{Generalized Linear Rules (GLR)}.
GAMs further relax the requirement of simple weighted-sum correlation and instead presume that the result can be characterized by a combination of arbitrary functions of each attribute.

\textbf{Generalized Additive Models (GAMs)}.
GAMs are useful for deciphering a part of assumptions of linear models (i.e. if the output y and a certain feature f without any feature interaction follow a Gaussian Distribution).
However, these linear model extensions result in a more complicated (i.e., more interactions) and less interpretable model.
Aside from the methods mentioned above, a Naive Bayes Classifier, which independently assesses the probability of classes for every feature, and K-Nearest Neighbors, which uses data point neighbours for prediction (regression or classification), are also considered intrinsically interpretable approaches.

Although intrinsically interpretable models can be applied to complex problems, they cannot aid in the understanding of machine learning techniques. As we progress to increasingly correlated feature sets and nonlinear problems, we must investigate another realm of explainable techniques.
In this area, we will examine strategies that merely leverage existing dataset information to provide system interpretation.
We'll divide them into three sections for the most part.
Visualization based approaches, Model Agnostic methods and Example-based approaches, respectively.

\subsubsection{Model Agnostic, Visualization based approaches}
\textbf{Partial Dependence Plot (PDP)}
PDPs (Partial Dependence Plots) are a well-known example of Model agnostic approaches. Model agnostic approaches are interpretation-based universal approaches that can create interpretation for any machine learning model.
These plots help represent partial impact of variables on the forecast. These are special tools for visualizing the non-linearity and other intricacies underlying the ML model. These global methodologies demonstrate the significance of features from the input data. However, they are 'partial,' as they can only show one or two features at a time. They also do not take into account feature interaction, which has an impact on projected outcomes.
\textbf{Individual Conditional Expectation (ICE) Plot}
The ICE plots give another visualization-based way for displaying each instance at one line. It demonstrates the effect of feature changes on instance's prediction.
For each feature, one line per instance means we need to plot n plots for n features, and each plot will contain 'z' lines to represent 'z' instances. This has more information than a single line overall, as in partial dependence charts. With each line per instance, ICE curves are intuitive to understand. The shortcoming of these curves is that we cannot detect the correlations' feature reliance on each instance.
\textbf{Accumulated Local Effects (ALE) Plot}
The ALE gives, on average, the features' influence on a machine learning model's prediction. These are more unbiased and faster to compute than PDPs, and their interpretation is more meaningful, but these graphs might become wobbly with larger intervals~\cite{Dream,aminBPM}. ALE charts, unlike PDPs, are not accompanied by ICE curves. These plots are more suitable for working with associated features.
\subsubsection{Model Agnostic, Feature Interaction Based}
\textbf{Feature Interaction}
In computation, features interact with one another. The key element of feature interaction is that the effects of individual features do not always add up to the combined effects. \cite{Islam2021ExplainableSurvey}, the interaction between two features is computed by calculating the difference of the partial dependency for two features and the sum of the partial dependency for each feature separately. This method has the disadvantage of being computationally expensive.
\textbf{Feature Importance}
The feature relevance increases when the feature values are permuted to disrupt the correspondence between the feature and outcome. If mistakes grow after rearranging the value of the feature, it emphasizes the relevance of the feature. The relevance of a feature gives a concise and global view of the ML model's behavior. The feature significance considers both the main feature effect and interaction which can be detrimental because feature interaction is included in importance of associated features. In  nutshell, the presence of feature interaction, significance of feature does not cause a performance fall. Furthermore, the training and test set is used for feature importance because it reflects iterative variation in the shuffled data set. It is also noteworthy that feature importance is included in the global methods.
\subsubsection{Model Agnostic, Global Surrogate}
This is an interpretable approximation model which is trained with the same data set to forecast similarly to a black-box model. We can understand the behavior of the black-box model by reading the surrogate model. We solve the black-box model by creating a new one. 
These models are also being chastised for providing a comprehensive grasp of the system, which cannot be classed as interpretation. The models discussed in Intrinsically interpretable models can be used to develop the global surrogate models.
\subsubsection{Model Agnostic, Local Surrogate}
This class of models are interpretable models employed in specific predictions made by black-box machine learning models. 

\textbf{LIME}.
Local interpretable model-agnostic explanations (LIME), proposed in the reference \cite{Ribeiro2016WhyClassifier}, is a local surrogate model. Rather than developing a global surrogate model, the LIME aims to train local surrogate models to interpret individual predictions. Another possibility is to create one or more local surrogate models. On selected subsets of the data, local surrogate models approximate the complicated model's predictions. In the case of a mortgage, several explainable models would be built for different sorts of mortgage applicants\cite{Ribeiro2016WhyClassifier}.


\textbf{Shapley Values}.
This is a game-based method to system prediction \cite{Kuhn1953ContributionsAM-28}. A prediction can be elaborated by supposing that the value of each feature instance represents a "player" in a game, and  the forecast is its "reward". The Shapley values, a mechanism derived from coalitions game theory, provide us with a scheme to fairly distribute the "payout" among the qualities. The Shapley value allows for opposing arguments and takes a long time to compute, yet it is susceptible to misinterpretation.

\textbf{SHAP}.
Some representative methods share some qualities, as mentioned above. Another newly proposed approach, SHapley Additive exPlanations (SHAP), proposed by Lundberg et al. \cite{Lundberg2017APredictions}, has taken advantage of a mixture of Shapley values from game theory \cite{Hart1989ShapleyValue}, the approach attempts to fairly assign the contributions. Rodriguez et al. employed this in their work. \cite{Rodriguez-Perez2020InterpretationPredictions} for explanation generation for pharmaceutical industry data set, by Hong et al. \cite{Hong2020RemainingReduction} used with DNN to predict health analytic tool.

\subsubsection {Propagation Based Methods}
The methods relying on local approximation offer another model to comprehend complex models locally. However the behavior in the local region may require a wide variety of features to generate comprehension, which complicates the process and leads to an uninterpretable system. The relevant approach is based on backpropagation and forwards propagation, and it quantifies the system that uses the features through output difference after perturbing the feature.

\textbf{Back-Propagation Based Methods}.
The approach is based on taking a derivative of the outcome using input features. One well-known effort is DEEPLift \cite{Shrikumar2017LearningDifferences}, which uses a back-propagation-based strategy to create an interpretation of crucial model aspects.

\textbf{Forward Propagation Based Methods}.
These methods leverage perturbations on the input to find the relevant feature for output \cite{Fong2017InterpretablePerturbation}. 

\subsubsection{Instance-based Explanations}
The approaches indicated above interpret the systems. However, they can be computationally costly for large setups, and the explanation generation time is insufficient. In such cases, it may be preferable to generate explanations for specific system examples rather than the overall one. This method does not necessarily explain the model but rather highlights its key elements. The approach generates answers to questions like, "What were the driving variables in the case of individual A?"

\textbf{Break Down}.
This category of models generates local explanations and is associated with the partial dependence algorithm using a step-by-step approach known as "Break Down." This is a greedy iterative technique for identifying influence-based characteristics on overall response. The game theory approach starts with a null team and gradually populates feature values, one by one, based on contributors' contributions. The amount of feature's contribution, in each iteration, from each feature, depends on the feature values of those already in the team, which is a disadvantage of feature in this strategy. Because of the greedy approach, this method is faster than the Shapley value method.


\textbf{Counterfactual}.
According to Islam et al. \cite{Islam2021ExplainableSurvey}, a counterfactual explanation characterizes a causal scenario: "If X had not occurred, Y would not have occurred." For example, "I would not have known the cause for the fame if I hadn't gone to this coffee shop."
The interpretations of this method are pretty straightforward, and these systems do not require access to data or models to anticipate. This method is also simple to build, but the constraint is to have numerous counterfactual representations for each system instance.

\textbf{Adversarial}.
Most approaches suggest reducing the gap between the antagonistic example and the instance to be controlled while adjusting the prognosis to the desired outcome of the system. This method allows for diagnosis.

\textbf{Feature Selection}.
Other proposed strategies include instance-based feature selection, which selects features depending on each instance. In this framework, a feature selector always attempts to maximize the quid pro quo information of the selected features and the response variables. This approach, however, is limited mainly to ad-hoc strategies. 
%
The techniques, as mentioned earlier, are all data drive techniques as all the results and explanations are generated from the available data. These techniques are solely dependent on the data set and/or the model for explanation generation. This is a major cause of a limitation on the explanation generation system. As to develop an explanation system understandable by the layman, we need to add knowledge to the system to generate more descriptive explanations.

\subsubsection{Neural Network-Specific Techniques}

Although we have covered Model agnostic local and global methods and visualization techniques for explanation purposes, studying the methods specific to NN is vital as in the NN feature learning is in the hidden layer of the network. To understand, we need the tools to interpret NN. We can also use gradient-based methods for interpretation which are computationally cost-efficient than model agnostic methods. 

\textbf{Gradient-Based approaches}.
Gradient Class Weight Activation Map (Grad-CAM) \cite{Selvaraju2017Grad-CAM:Localization} is a visualization based approach. It is based on the gradient of the neural network. It assigns the relevance score to each node for the relevance of the score to the decision. Many Studies have used Grad-Cam and have reported results including \cite{Choi2020InterpretingGrad-CAM} but not limited to. Other approaches include Smooth-Grad\cite{Smilkov2017SmoothGrad:Noise}, DecovNet \cite{Noh2015LearningSegmentation}, Vanilla Gradient \cite{Simonyan2014DeepMaps}, Integrated gradients \cite{Sundararajan2017AxiomaticNetworks}.

\textbf{Graph Neural Network (GNN)}.
Graph Neural Networks (GNNs) are vital tool for machine learning on graphs. We have explanation tools available for GNN. One to mention is GNNExplainer \cite{Zhou2020GraphApplications}.

\subsubsection{Knowledge-Based Techniques}

So far, the methodologies we've looked at have only focused on using data set metadata to explain.
The following simple step towards high fidelity explanation is to incorporate information (e.g., expert domain knowledge and/or corpus knowledge to explain the system).
The information can help to extract the structure from the ML system to deliver the explanations. 
The explanation section can be expanded to create more useful information based on domain knowledge or expert knowledge. This category is being used to investigate the knowledge-based methodologies to utilize this information briefly. How might they bring value to the development of explanations?
We can find studies in the knowledge-based explanation systems. They are contributing expert domain knowledge or corpora to the system. The overall discussion of the methods is available \cite{tiddi2020knowledge}. In this section, we will discuss a few scenarios. 

\textbf{Quantitative Testing Concept Additive Vectors (TCAV)}.
One such study is by Kim et al.\cite{Kim2017InterpretabilityTCAV} about the NN human-understandable interpretation. The proposed approach is TCAV, a post-doc, Model agnostic approach. This approach provides human-understandable explanations. This is a significant study of knowledge-based systems.

\textbf{Knowledge Infused Learning (K-IL)}.
K-IL is another proposed approach in this study~\cite{Kursuncu2019KnowledgeLearning}. This approach uses knowledge graphs to generate an understanding of the system. This study claims that knowledge graphs are way forward to the explanation generations. 

With the discussion of knowledge-based approaches, we wrap up this section and have summarized the selected approaches in the table\ref{tab:existing_approaches}. From this table, we can see that most of the generated interpretations are not humanly understandable. We need a domain expert or AI layman to understand the explanations generated. Whereas table  \ref{tab:xai_evaluation} highlights that explanation evaluation systems are not widely developed. Few techniques have quantitative or qualitative evaluations system. This lack of existing approaches is a way forward to establish an explanation system for human understanding and relevant evaluations. These methods can help the experts understand how the system achieved specific output from the given feature set, but these are not enough to understand why a certain decision was derived from the data set. To make it more human-understandable, we need to provide interactive system frameworks for interactions \cite{Miller2019ExplanationSciences}. 

\subsection{XAI for Time Series Data}
In the literature, we can find surveys about the existing techniques of XAI. 
Most of the studies for XAI revolves around Image data sets as our focus is on bank transaction Time series data in this case. We want to understand the impact of XAI on time series data in the literature. 
There are few surveys available for the overview of XAI approaches for time series data.
A recent survey on time series details by Rotaj et al. \cite{Rojat2021ExplainableSurvey}. The main categories for the time series data set as mentioned in the study by Rotaj et al. are explanations for Convolution Neural Network , explanations for Recurrent Neural Network and Data mining for explanation generation based approaches \cite{Rojat2021ExplainableSurvey}.  

\subsubsection{Evaluations of Explanations}
The studies by Schlegel et al.  \cite{Schlegel2019TowardsSeries}, \cite{Schlegel2020AnTasks} have used various XAI based methods SHAP \cite{Hart1989ShapleyValue} LIME \cite{Ribeiro2016WhyClassifier} LORE \cite{Guidotti2018AModels} are to name a few, for explanation generations.
According to the evaluation and analysis in the studies mentioned above, SHAP explanations are more robust, but other approaches, although model agnostic, can perform better with certain architectures. 
One of the significant downsides for XAI based method is the non-verifiable results. 
Few approaches to discuss the application of XAI in the time series data are presented. 

\subsubsection{Prototype-based Approach}
The Proposed prototype-based approach by Gee et al. \cite{Gee2019ExplainingPrototypes} is an example of the deep classification system for time series data. In this system, an encoder-decoder based prototype is designed, and that prototype was used with the system to generate the explanations for the time series dataset.
\subsubsection{Theory Driven Approach}
The approach proposed by Wang et al. \cite{Wang2019DesigningAI} is theory-driven. This is user-centric that means for each user, it evaluates the scope. Its bridges the algorithm with human decision making theories

\textbf{XAI for Fintech}.
When it comes to XAI, Fintech is an important area for the application of XAI. Few to mention studies working on XAI for fintech are \cite{Fritz-Morgenthal2021FinancialAI}, \cite{Bracke2019MachineAnalysis}. They have explored XAI techniques for the application in this domain and are identifying the practical advice for responsible XAI systems in production. 

\subsection{Interactive Tool for Insights}
XAI has gained fame due to its ability to provide insights. Along with explainability, knowledge-based systems are proposed to take humans in the loop. The human in the loop approach can be beneficial to solve computationally complex problems. 
To add, humans-in-the-loop Interactive Machine learning platforms are getting the limelight. One of the lists of commercially available tools for visualisation and interaction is available online \cite{KijkoTheNeptune.ai}.  There are a few tools we will discuss next. These can analyse system behaviour interactively.  

\subsubsection{What-if Tool}
Google has proposed What-if, a machine learning-based diagnostic tool \cite{Wexler2020TheModels}. It is an interactive tool for exploring and probing machine learning models. The limitation of this tool is still the black box. This black box nature is a limitation of the explainable nature of the system. The associated metrics of XAI systems are not measurable as well, and they cannot ensure fairness. The proposed approach has different data sorting capabilities based on mathematical fairness measures to ensure fairness in the system  \cite{Weinberger2021PlayingFairness}.

\subsubsection{Explanatory Interactive Machine Learning (CAIPI)}
Explanatory Interactive Machine Learning proposed CAIPI for feedback to users by Teso et al. \cite{Teso2019ExplanatoryLearning}. This is a model agnostic tool that provides a framework for active learning. This tool can learn by querying the users. With the active learning part, this tool can give some information on learnt knowledge from machines. But this system still has uncertainties involved as it depends on how ML developed system parameters can interact with the systems. 

\subsubsection{Contextual and Semantic Explanations (CaSE)}
Contextual and Semantic Explanations (CaSE) architecture proposed by Keifer et al. \cite{Kiefer2022CaSE:Knowledge} is a text classification explanation tool. It combines the local explanation model with the semantic and contextual meanings to explain the system. 
So it mainly comprises three components.
\begin{itemize}
    \item Classification algorithm
    \item Explanation algorithm (local)
    \item Semantic approach (topic modelling in text mining)
\end{itemize}

Semantic interrogation ability is to map the human explanations to understand as the human thinking process is the coherent ability to question. How to ensure the validity and accuracy of the explanation, it needs to evaluate the performance compared to other similar systems.
\subsection{Explainability Toolbox}
The study by Bracke et al. \cite{Bracke2019StaffAnalysis} proposed a five step explainability framework. The proposed approach is a combination of several features of existing methods. They begin with an instance based system and proceed to generate global explanations, similar to a Feature Importance. However, feature interactions can be captured. The visualisation step is similar to PDP but are also based on Shapley values similar to instance based. This allowed non-linearity, identical to PDPs, yet capture feature interactions and aims to capture model behaviour on a granular level.

\par
\subsubsection{Human-in-the-Loop}
Another study by Holzinger et al. \cite{Holzinger2018FromAI} studies the impacts of humans in the loop. This study emphasises that adding a human in the loop can make the computationally challenging problem easy to solve. They also propose that humans in the loop can help to get better explanations from the system. With the use of gamification, humans and machine can form a coalition. 
These studies are fascinating, and we can explore avenues of research in these studies, but the evaluation of such a task is critical, as mentioned in this study \cite{Boukhelifa2020ChallengesSystems}, \cite{EhsanHuman-centeredApproach}.

\begin{table*}[htbp]
\vspace{-40pt}
\caption{Summar of the selected XAI based approaches and provide a summary of the major attributes of the system. Column I, II Names and Citation, column III states the approach is ante-hoc or post-hoc, column IV mention methodology, column V indicates the target audience, column VI provides the information about explanation types column VII provides the scope information.}
\begin{center}
\resizebox{0.85\textwidth}{!}{%
\begin{tabular}{|c|c|c|c|c|c|c|}
\hline
{\textbf{ \makecell{Selected XAI \\Approaches } }} &  {\textbf{Citation }}&\textbf{\textit{\makecell{Ante-hoc\\/ Post-hoc}}}& \textbf{\textit{\makecell{Explanations\\Methodology}}}  & \textbf{\textit{\makecell{Target \\Audiance}}} & \textbf{\textit{\makecell{Explanation \\ Type}}} & \textbf{\textit{\makecell{Scope}}}
\\
\hline
TREPAN & \cite{Craven1995ExtractingNetworks} & Post-hoc & \makecell{Surrogate Model \\(Decision Tree)} & \makecell{Domain\\ Experts} & Approx.& Global \\ \cline{1-7}
Deep Red & \cite{Zilke2016DeepREDNetworks}  & Post-hoc & \makecell{Surrogate Model \\(Decision Tree) } & \makecell{Domain\\ Experts} & Approx.& Global \\ \cline{1-7}
Distilling NN&  \cite{Frosst2017DistillingTree} & Post-hoc & \makecell{Surrogate Model \\(Decision Tree) } & \makecell{Domain\\ Experts} & Approx.& Global \\ \cline{1-7}
PFI &\cite{Nguyen2016SynthesizingNetworks} & Post-hoc & \makecell{Feature \\Attribution } & \makecell{Domain\\ Experts} & Approx.& Global \\ \cline{1-7}
DGN-AM &\cite{Nguyen2016SynthesizingNetworks} & Post-hoc & \makecell{Surrogate Model \\(Decision Tree) } & \makecell{Domain\\ Experts} & Approx.& Global \\ \cline{1-7}
\makecell{Interpretable\\ CNN} &\cite{Zhang2017InterpretableNetworks} & Ante-hoc & \makecell{Visualization \\ Mask} & \makecell{Domain \\Expert} & Approx. & Global\\ \cline{1-7}
LIME&\cite{Ribeiro2016WhyClassifier} &  Post-hoc & \makecell{Surrogate Model} & \makecell{Domain\\ Experts} & Approx.& Local\\ \cline{1-7}
LORE &\cite{Guidotti2018LocalSystems}& Post-hoc & Decision Tree  & \makecell{AI\\ Layman } & Approx. & Local\\ \cline{1-7}
\makecell{Vanilla \\Gradient} &\cite{Baehrens2010HowUller} &  Post-hoc & \makecell{Visualization} & \makecell{AI\\ 
Layman} & Approx.& Local\\ \cline{1-7}
Grad-CAM &\cite{Selvaraju2017Grad-CAM:Localization}& Post-hoc & Visualization  & \makecell{AI\\ Layman} & Approx. & Local \\ \cline{1-7}
DeepLift & \cite{Shrikumar2017LearningDifferences} & Post-hoc & \makecell{Feature \\ Attribution} & \makecell{Domain\\ Experts} & Approx. & Local\\ \cline{1-7}
SHAP&\cite{Lundberg2017APredictions} &  Post-hoc & \makecell{Feature\\ Attribution} & \makecell{AI\\ Layman} & Approx.& Local\\ \cline{1-7}
\makecell{MMD\\- critic} &\cite{Kim2016ExamplesInterpretability}&  --- &   \makecell{Prototype \\ \& Criticism} &\makecell{Domain\\ Experts} & \makecell{Instance \\based} & Specific \\ \cline{1-7}
Unconditional Counterfactual &\cite{Wachter2017CounterfactualGDPR} &  Post-hoc & \makecell{Data \\ Instance}  & \makecell{Domain\\ Experts} & \makecell{Instance \\ based}  & Specific\\ \cline{1-7}
PDL & \cite{Gee2019ExplainingPrototypes} &  Ante-hoc & Prototype  & \makecell{Developer} & \makecell{Case \\ based} & Global \\ \cline{1-7}
Doctor XAI & \cite{Panigutti2020AnExplanations} & Post-hoc & \makecell{Ontology\\based} & \makecell{Domain\\ Experts} & Approx. & Global\\ \cline{1-7}
 TCAV & \cite{Kim2018InterpretabilityTCAV} & Post-hoc &\makecell{Concept \\Importance} &\makecell{Domain\\ Experts} & Approx. & Global\\ \cline{1-7}
K-IL & \cite{Kursuncu2019KnowledgeLearning} & Post-hoc & \makecell{Knowledge \\Graph}&  \makecell{Domain\\ Experts} & Approx. & Local\\ \cline{1-7}
IBD & \cite{Zhou2018InterpretableExplanation}& Post-hoc & \makecell{Visualization \\ \& Concept \\Importance} & \makecell{AI\\ Experts} & Approx.& Global\\ \cline{1-7}
RuleRec &\cite{Ma2019JointlyGraph} & Post-hoc & \makecell{Decision \\Rule} & \makecell{AI\\ Layman} & Approx.& Local\\ \cline{1-7}
\makecell{propositional\\knowledge} & \cite{Labaf2017PropositionalKnowledge}& Post-hoc & \makecell{Knowledge  \\ Graph} & \makecell{Experts} & Approx.& Global\\ \cline{1-7}
\hline
\multicolumn{6}{l}{$^{\mathrm{a}}$ Approx. Approximate}
\end{tabular}
}
\label{tab:existing_approaches}
\end{center}
\end{table*}
\pagebreak

\begin{table*}[htbp]
\vspace{-41pt}
\caption{Summary of selected XAI based approaches and compare performance and evaluations. Column I (names of the approaches), column II citation, column III highlights is there any explanations evaluations provided?, column IV black box model explanations. AGN means Model Agnostic, column V which models are used to generate the explanations, column VI indicate the approach is data driven or knowledge driven, column VII highlights the behaviour of the explanation system.}
\begin{center}
\resizebox{0.85\textwidth}{!}{%
\begin{tabular}{|c|c|c|c|c|c|c|}
\hline
{\textbf{ \makecell{Selected XAI \\Approaches } }} &  {\textbf{Citation}}& \textbf{\textit{\makecell{Explanation \\Evaluation}}} & \textbf{\textit{\makecell{Black \\Box}}} & \textbf{\textit{\makecell{Data Driven/ \\Knowledge }}} & \textbf{\textit{\makecell{Behaviour/ \\Performance}}}
\\
\hline
TREPAN & \cite{Craven1995ExtractingNetworks} & No & NN & Data-Driven & \makecell{Tree\\ Extraction} \\ \cline{1-6}
Deep Red & \cite{Zilke2016DeepREDNetworks}  & No & NN & Data Driven & \makecell{Rule\\ Extraction} \\ \cline{1-6}
Distilling NN &  \cite{Frosst2017DistillingTree} & No & DNN & Data Driven & \makecell{Extract\\ Model} \\ \cline{1-6}
PFI &\cite{Nguyen2016SynthesizingNetworks} & No & AGN & Data Driven & \makecell{Feature \\ Importance} \\ \cline{1-6}
DGN-AM &\cite{Nguyen2016SynthesizingNetworks} & No & DNN & Data Driven & \makecell{Extract\\ Model} \\ \cline{1-6}
\makecell{Interpretable\\ CNN} &\cite{Zhang2017InterpretableNetworks} & No & CNN & Data Driven & \makecell{Model\\ Design} \\ \cline{1-6}

LIME&\cite{Ribeiro2016WhyClassifier} &No & AGN & Data Driven & \makecell{Local \\Approx.}\\ \cline{1-6}
LORE &\cite{Guidotti2018LocalSystems} & No & AGN & Data Driven & \makecell{Rule \\Based.} \\ \cline{1-6}
\makecell{Vanilla \\Gradient} &\cite{Baehrens2010HowUller} & No & CNN & Data Driven & \makecell{Propagation \\ based}\\ \cline{1-6}
Grad-CAM &\cite{Selvaraju2017Grad-CAM:Localization}&No & CNN & Data Driven & \makecell{Propagation \\Based.} \\ \cline{1-6}
DeepLift & \cite{Shrikumar2017LearningDifferences} & No & AGN & Data- Driven & \makecell{Feature \\Importance}\\ \cline{1-6}
SHAP&\cite{Lundberg2017APredictions}  & Quantitative  & AGN & Data Driven & \makecell{Feature \\ Importance}\\ \cline{1-6}
\makecell{MMD\\- critic} &\cite{Kim2016ExamplesInterpretability}&  Quantitative & AGN & Data Driven &\makecell{Instance \\based}  \\\cline{1-6}
\makecell{Unconditional \\Counterfactual} &\cite{Wachter2017CounterfactualGDPR} &  Quantitative & CNN &  Data Driven & Counterfactuals\\ \cline{1-6}
PDL & \cite{Gee2019ExplainingPrototypes} & Qualitative & DNN & Data Driven & \makecell{Prototype\\ Based} \\ \cline{1-6}
 \makecell{Doctor\\ XAI} & \cite{Panigutti2020AnExplanations} & No & AGN & Data Driven & Rule based\\\cline{1-6}
 TCAV & \cite{Kim2018InterpretabilityTCAV} & Quantitative & DNN & \makecell{Knowledge \\aware}& \makecell{Human\\in loop} \\\cline{1-6}
K-IL & \cite{Kursuncu2019KnowledgeLearning} & No & CNN& \makecell{Knowledge \\aware}& \makecell{External \\ Knowledge}\\\cline{1-6}
IBD & \cite{Zhou2018InterpretableExplanation}& No & CNN & \makecell{Knowledge\\aware} & \makecell{Corpus}\\ \cline{1-6}
RuleRec &\cite{Ma2019JointlyGraph} & No & --- & \makecell{Knowledge\\based} & \makecell{Internal \\Knowledge}\\ \cline{1-6}
\makecell{propositional\\knowledge} & \cite{Labaf2017PropositionalKnowledge}& No & AGN & \makecell{Knowledge\\aware} & \makecell{External \\knowledge}\\ \cline{1-6}
\multicolumn{6}{l}{$^{\mathrm{a}}$ AGN: Agnostic, CNN: Convolution Neural Networks, NN: Neural networks, DNN:Deep Neural Networks}
\end{tabular}
}
\label{tab:xai_evaluation}
\end{center}
\end{table*}

\pagebreak

\subsection{Limitations}

The well established limitation of the existing systems are that the generated explanations are for limited user base and cannot provide enough insights for the layman.
These explanations also doesn't have well established evaluation mechanism. Which are a research pathway to explore. We need to have a evaluation mechanism that can provide us the insights to the system and can establish the grounds that system generated explanations can qualify the associated metrics of trustworthiness, confidence and reproducible of the results. \par
According to Tim Miller \cite{Miller2019ExplanationSciences} we need not to reinvent the wheel for understanding of the systems. To generate better explanations we need to understand the human thinking processes as these are already established studies in social sciences and psychology; using those theories we can build the models close to human understanding~\cite{Beheshti2020Personality2vec}. 

 
\section{Proposed Model} \label{chap:Methodology}

\begin{figure}
     \centering
     \includegraphics[width=15.5cm, scale=2.5]{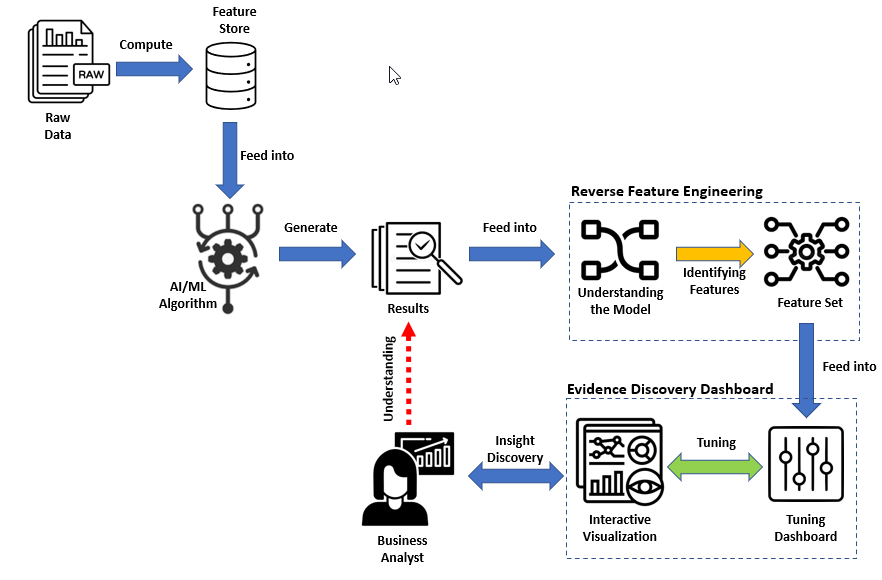}
     \caption{Overview of the architecture of the proposed evidence-based pipeline. This figure describes the information flow in the pipeline and highlights the major components presented in the study}.
     \label{fig:methodology}
 \end{figure}

This section walks through the proposed evidence-based pipeline to achieve a better understanding of machine learning algorithms. The proposed approach is generic and model agnostic. We can implement this pipeline for the evaluation of any machine learning model. For the scope of this study, we implemented this pipeline on the bank transaction classification system. After reading this section, the reader will understand the proposed model and its application. 
To solve the explainability issues associated with the machine learning algorithms. We are submitting a prototype framework to explain with evidence established with visualization and feature importance.  
The fundamental purpose of the proposed pipeline is to help explain the results of machine learning algorithms to data scientists and domain experts by finding the appropriate shreds of evidence from the available data~\cite{CoreDB,CoreKG}. This system can help gain better insights into ML model predictions with the help of visualization and interactive tuneable dashboard to maintain the intelligent design and gain insights on the particular decision. This interaction can strengthen the understanding of prediction results and assist in analyzing the feature contribution. It can be a helpful tool to guide the data scientist to comprehend the prediction's evolution better. The proposed system can assist to reverse engineer the feature vectors based on the importance of the features obtained with the evidence from the system~\cite{Assessment2Vec}. This system automates the insights of the ML algorithms for the Credit Bank Transaction Classification of the customer. The architecture of the proposed pipeline is shown in the Figure \ref{fig:methodology}. In the following sections, we will provide detail of each section of the pipeline. 

\subsection{Data Moderation for Feature Engineering and Evidence Discovery Dashboard}

This study aims to support the finance officer, the data scientist and other business stakeholders with the platform to analyze the feature importance and identify the relevant transactions based on the similarity provided in this evidence-based system. With this pipeline, the system audience will have access to a portal that can provide insights into the system. With this portal, we can use the details of the transaction to understand the transaction classification. 

\subsubsection{Bank Classification Raw Data Moderation}

In any sector, the importance of data curation is a vital step. Good quality data sources and organization is essential to improve the performance of the system~\cite{DataSynapse,curationAPI,iStory}.  We need curated information in a way that can help to improve the information extraction from the framework.
The business document such as bank statement, client loan history, loan applications, income, and more are recognized as first-class citizens in a financial loan application setting, allowing the process to focus on the essential informational entities. Using appropriate data curation approaches to such business documents, we may identify the hidden insight and information described in process-related objects. Apart from the amount of the transaction, a bank transaction might contain information on the type of device used, GPS location, transaction time and  description. As illustrated in Figure \ref{fig:methodology}, the initial step of our pipeline focuses on moderation of raw bank transactions into relevant and meaningful data for the system.

\subsubsection{Acquisition of Raw Data}

We define a business document as an independant data object with a unique identity that can be described by an attribute vector Object\textsubscript{id}, Object\textsubscript{category} and Object\textsubscript{schema} where, $Object\textsubscript{id}$ is a primary attribute to represent the unique identity of business document; Object\textsubscript{category} and  Object\textsubscript{schema} are mandating elements to represent the type and metadata of the business object respectively.
A bank transaction is an example of a business document. Various characteristics are extracted during the extraction step, including:

\begin{enumerate}
    \item Schema features of a transaction may contain time, location, source, and description;
    \item Lexical features of descriptive part of a transaction can include; keyword, abbreviation, and bank-specific identifiers like; ATM code or bank code.
\item Natural Language Processing (NLP) of a transaction include; Part-Of-Speech tagging, Named Entity Identification.

\item Location information in the transaction.
\end{enumerate}

At this phase, domain-specific entities mentioned in the list in section \ref{chap:motivating}.2.1 will enrich the data to identify the potential named entities to add more information to raw data for the next task.

\subsubsection{Feature Store}

\textbf{Feature Engineering}

Feature Engineering is the process of extracting characteristics from raw data for prediction models to fully comprehend the dataset and perform effectively on previously unknown data. Feature engineering isn't a one-size-fits-all approach. This vital step determines the success or failure of the models~\cite{Khadivizand2020,FeatureAlireza,FeatureNRez}.
Each situation has a distinct representation of datasets for machine learning algorithms. Certain words may make a difference when it comes to banking. The contextual information helps to extract meaningful information. 
The curated raw data is the source of truth for the system. The carefully curated data will help plan to improve the performance and have a foundation for the analysis of the system. The curated data is used to perform feature engineering. The feature selection approach helps to get the engineered features from the available feature set. These selected featured are placed in the feature store for the respective model. The feature vector selection for feature store production is essential in the ML algorithm development lifeline. 
Numerous feature engineering tools are available for feature development. This pipeline stage can be automated with the relevant feature engineering tool to extract features and store them in the feature store for the model. \href{https://github.com/alteryx/featuretools}{Featuretools}\footnote{https://github.com/alteryx/featuretools}, Autofeat TSFresh are to name a few. With the help of these tools, the curated data is input to the feature engineering tool, and it extracts features from the data and performs transformations. After transformation, features are selected from the list with the output parameter.
Feature selection techniques are used to select the appropriate feature from the provided list. Feature selection techniques are generally divided into supervised and unsupervised methods. Unsupervised and supervised techniques are classified as wrapper, filter, and intrinsic methods. Feature selection approaches based on feature filtration assess the dependency of input variables to select features using statistical measures.

\subsubsection{Intelligent Algorithm and Results}
Machine learning algorithms are of various types, and multiple algorithms are used to solve numerous problems. 
In our study, the focus is the multi-class classification problem of bank transaction artifacts. There are three known ways to solve multi-class classification problems.

\begin{itemize}
    \item Assign a binary variable to each class and forecast individually. Further, logistic regression, decision tree algorithms can be employed.
    \item Naive Bayes algorithm, Neural Networks and Support Vector Machines (SVM) can be a choice to solve multi-class problem.
    \item Multi-layer modelling consists of multi level statistically parameterized model.
\end{itemize}

\begin{figure}
    \centering
    \includegraphics[width = \textwidth]{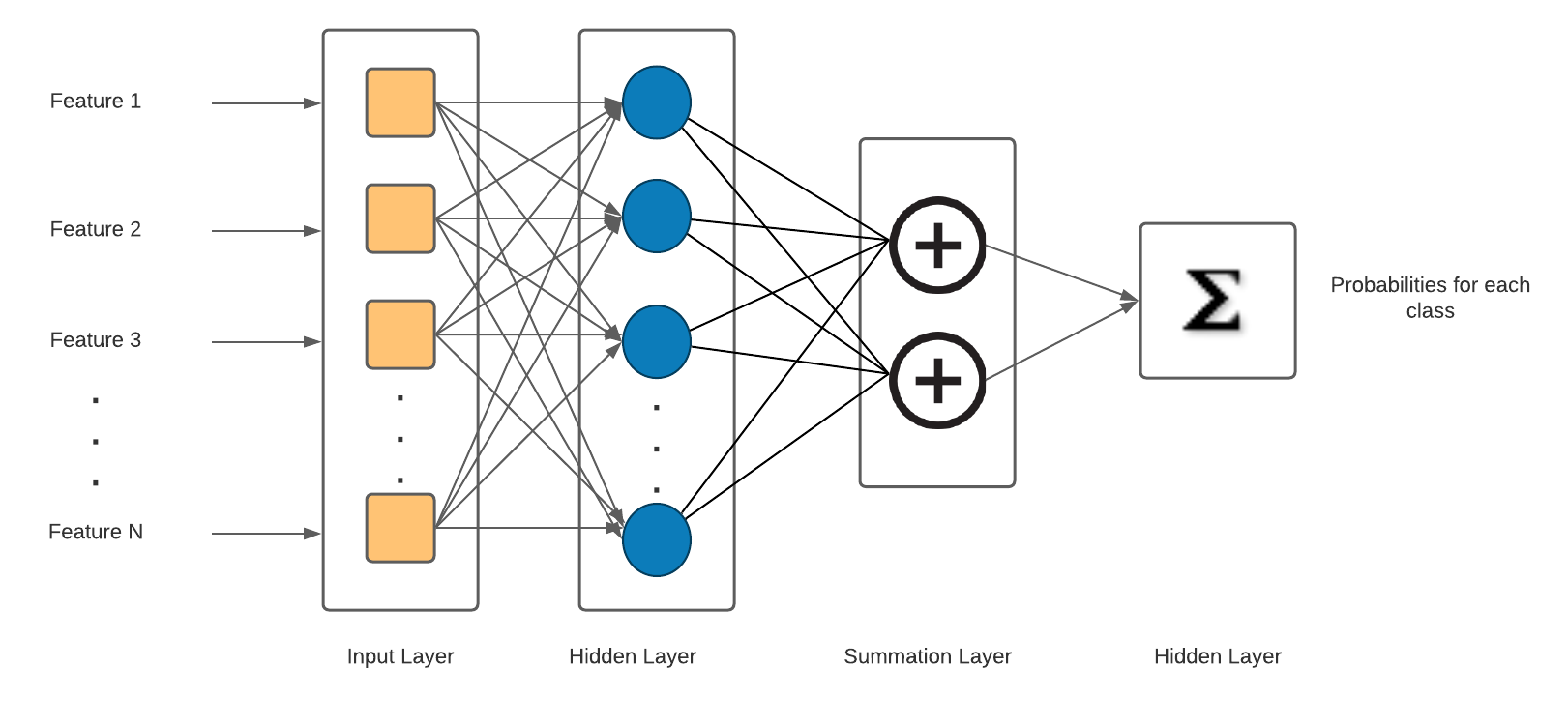}
    \caption{Probabilistic Neural Network Architecture used in the system for the better understanding of the input and output of the network, as we need them for the development of the evidence-based system in the pipeline}
    \label{fig:PNN}
\end{figure}

In our case, we are using a Probabilistic Neural Network (PNN) to classify the multi-class scenario of bank transaction classification. The PNN architecture is shown in Figure \ref{fig:PNN} to provide the understanding of the system. The final result is assigned based on the highest probability of all classes. 

The output generated by the system is probabilistic, and the sum of all the results is unity. For the final classification, we assign the highest probability of all the classes as our scenario has only credit transactions, so the possibility of having similar transactions is very high. 
The transactions are close enough, and two classes can have a close predicted value where we have to choose one class as a final answer. The scenarios can be very similar and complicated to understand the classification, so we need to add some additional layers for the transparency of the PNN classification. Few transaction classification probabilities are  available in the Table \ref{tab:class}
We have PNN predictions, but we cannot answer the questions of what and how of the system as mentioned in detail in the motivating scenario section \ref{chap:motivating}. 
\begin{table}
    \centering
     \caption{Sample of probabilities assigned to the classified transactions using PNN with predicted and actual models for the transaction}
    \begin{tabular}{|c|c|c|c|c|c|c|c|}
    \toprule
         Sr. No & \makecell{ Actual \\Classification} & \makecell{Predicted \\Classification} & Funding & Invoicing & Cash& Cheque & Other  \\
         \midrule
         1 & Invoicing	& Invoicing &0.14 & 0.328 & 0.325 &	0.074	&0.141 \\ 
2& Funding & Funding& 0.41 & 0.023 & 0.273
& 0.005& 0.34
\\
3 &Funding &Funding&\textbf{ 0.551	}&0.062&	0.021	&0.35&	0.185 \\
4 &Invoicing & Cash &0.141	&0.314	&\textbf{0.329}	&0.075&	0.142\\
5 &Invoicing & Cash &0.070&	0.133&	\textbf{0.577}&	0.021&	0.199\\
6&Funding & Funding &\textbf{0.536}&	0.005&	0.201&	0.125&	0.133\\
7& Funding &Other  &0.214&	0.063	&0.273&	0.005&	\textbf{0.445}\\
8&Invoicing &Cheque&0.056&	0.203&	0.217&\textbf{	0.498}&	0.026\\
9& Invoicing& Cash&0.155	&0.306&	0.\textbf{312}&	0.147&	0.080\\


\bottomrule
    \end{tabular}
   
    \label{tab:class}
   
\end{table}

\section{Reverse Feature Engineering}
The results of the machine learning algorithm and the input feature vector of the transactions are input to this next stage of the pipeline, reverse feature engineering. This component provides the feature score for the input features, which are feedback to the wrapper model for feature selection. It helps to understand the importance of features. We can analyze which attributes contribute to the output based on the score calculated by the feature importance and  by tweaking the system's input and changing the outcome. This interactive system helps to understand the significance. Based on this interaction, a data scientist can select features appropriately to improve the performance of design and can reduce the computation course by providing an optimal list of attributes. This proposed component is agnostic as this approach can be used with any model results and input feature vectors. The main contribution of this approach is we can analyze various transactions to understand the impact of the features on the output.  
These approaches are stochastic, so results may vary based on the provided dataset. 
\section{Evidence Discovery Dashboard}
Insights-A form of the post-hoc model understanding the existing dataset known as evidence. 
To perceive the model, the data scientist can study the retrieved results of the feature and analyze them with historical data to understand the system performance. With the evidence tuneable dashboard, we can analyze the situation. The proposed dashboard helps to grasp the standing of the current instance of the transaction concerning the feature(s) on other similar transaction features and results. The proposed dashboard is a web-based portal to support domain experts and other business stakeholders for convenience and availability. The tuneable feature component provides the ability to alter the instance attributes and reflect the change in the behaviour of the ML prediction. 

\subsection{Tuning Dashboard}
\begin{figure}
    \centering
    \includegraphics[width=0.95\textwidth]{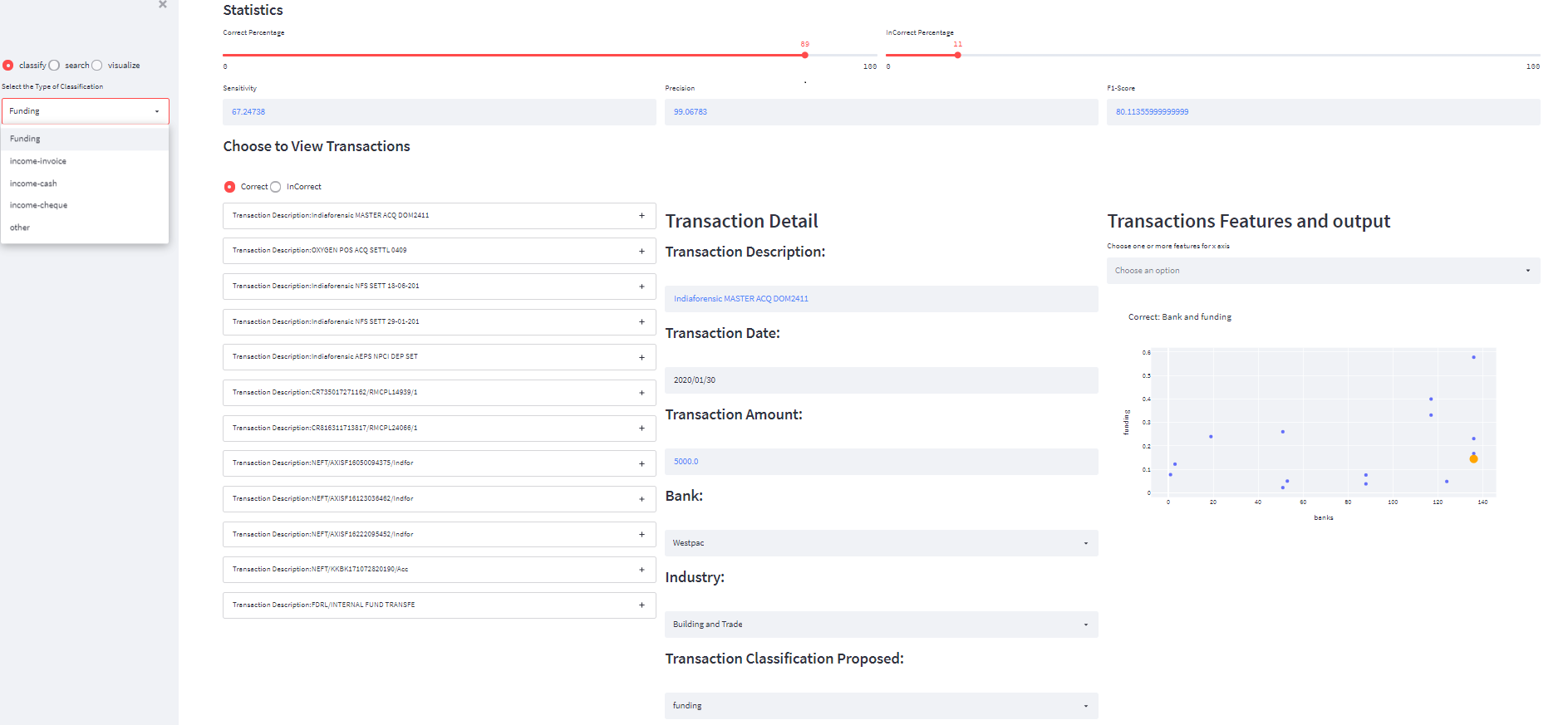}
    \caption{The snapshot of the developed interactive dashboard. The visible panel is the classification based filtering panel to select the classification type and show the results after segregating to correct and incorrect. The further detail feature in the panel helps to expand the transaction for detail understanding and tweaking the data to change the prediction response with change of feature vector. The last panel helps to understand the standing of the transaction in the existing available data with respect to the selected feature.}
    \label{fig:classification}
\end{figure}

\begin{figure}
    \centering
    \includegraphics[width=0.9\textwidth]{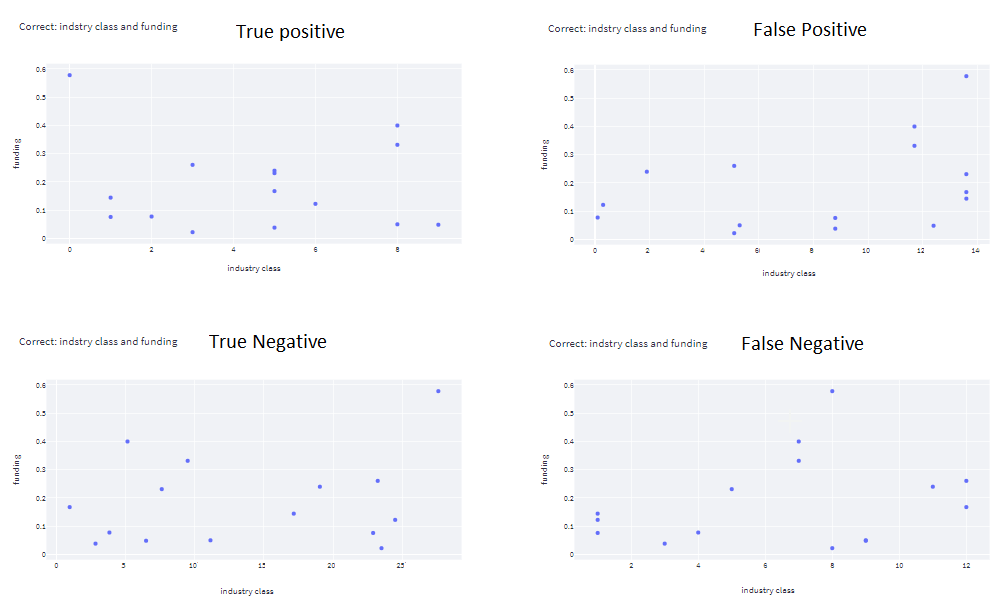}
    \caption{This is the screenshot of the visualization component of the developed interaction board. This section help to visualize the data instances in various way. In this snapshot it shows the standing of the data instances for the funding category in terms of TP, FN, TN and FP. The data entries are reduces for understanding purposes.}
    \label{fig:Visualization}
\end{figure}

\begin{figure}
    \centering
    \includegraphics[width=\textwidth ,height=0.7\textheight]{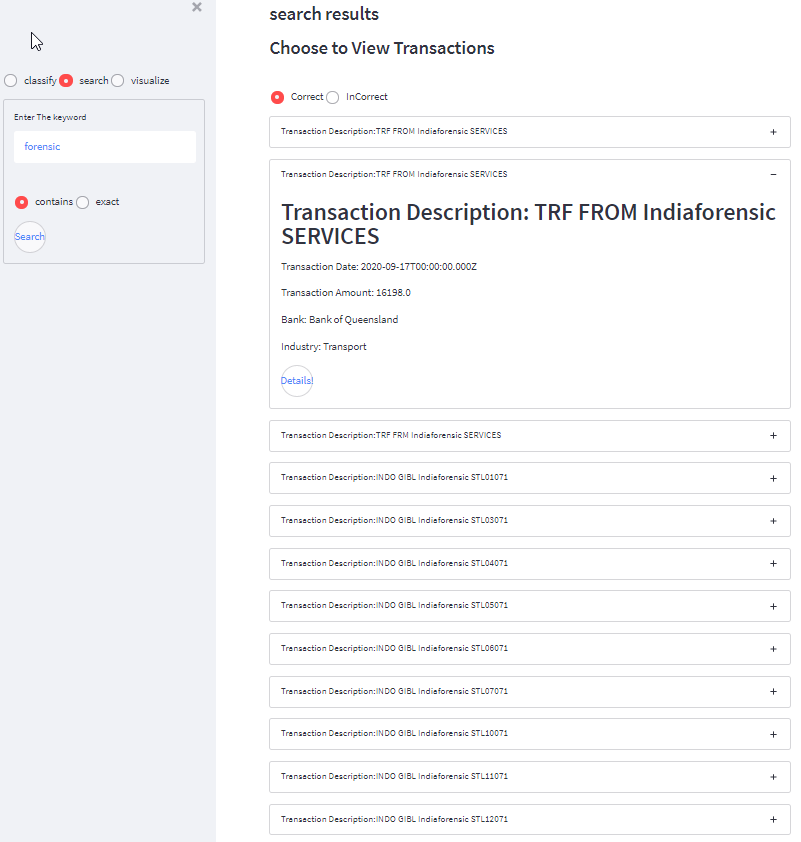}
    \caption{The snapshot of the search support panel of the developed dashboard. This dashboard provide the provision to search the transactions contains or have exact match with the search term available. The selected results are split on the basic of correct and incorrect classification for better understanding.}
    \label{fig:search}
\end{figure}

As mentioned above, this dashboard supports input feature tuning capability to reflect the system's output and visualize predicted behaviour. 
Available Tuning Options of the dashboard are listed below. 
\begin{itemize}
    \item Classification based filtering and segregation of correct and incorrectly classified transactions in the classification. 
    \item Search support for Transaction Description and filter based on correct and incorrectly classified.
    \item Overall Visualization of System to understand the standing of instances based on Actual data and Predictions. 
\end{itemize}
The dashboard snippets of the system are added as images, We have replaced the actual data with the publically available Kaggle\footnote{https://www.kaggle.com/apoorvwatsky/bank-transaction-data} dataset transaction description to avoid customer privacy breach. The Figure \ref{fig:classification} shows that we can choose the classification from the given system category and calculate precision, sensitivity, and F1-score. Then filter the transaction based on correctly classified and incorrectly classified. With this front end, we can check the detail of any transaction and visualize the standing of the transaction with any input feature for understanding. As shown in the figure \ref{fig:search}, the feature search supports the ability to search specific transaction contains or have an exact match for the search terms. With this capability, the user can get further information about the relevant transaction. Similarly, we can generate feature-based insights for each one of the transactions. The visualization platform provides a detailed view of the transaction results, as shown in the Figure \ref{fig:Visualization}. This dashboard helps to improve the trust and fairness in the model predictions.

\subsection{Interactive Visualization} 
With the help of interactive visualization, one can observe the behaviour of the transactions and the predictions based on the input. This helps to understand the system contribution. 
Interactive visualization is based on.

 \begin{itemize}
     \item Filter on Classification basis
     \item Filter on Application basis 
     \item Filter on all the input feature basis
     \item Tweak the input and recalculate the results of the ML algorithm.
     \item Comparison of the final result with the other probabilities to identify the similarity.
 \end{itemize}

\subsection{Algorithm for Evidence-based Visualization Dashboard}
\begin{algorithm}
\caption{Algorithm for the Evidence-based Approach}\label{alg:evidence}
\begin{algorithmic}[1]
\REQUIRE $Business-Artifact$
\REQUIRE $Feature Vector$
\REQUIRE $Decision$
\ENSURE $Insights$
\STATE $Retrieving-Schema \gets Business-Artifact-scheme$
\STATE $Feature-Data \gets Feature-vector$
\WHILE{$Feature-Data\geq0$}
\STATE $Read-Feature-Vector$
\STATE $Retrieve-Decision$
\STATE $Join-Features-with-Decisions$
\STATE $Decision-Evaluations-calculation$
\ENDWHILE
\IF{$Decision-category$}
\STATE $Load-category-filtered-results()$
\STATE $Load-category-specific-calculations()$
\IF{$Artifact-detail \geq 0$}
\STATE $Load-specific-artifact$
\STATE $Visualization-of-artifact-with-similar-attributes$
\STATE $Insights$
\ENDIF
\ENDIF

\IF{$Search-category$}
        \STATE $Enter-search-term()$
        \STATE $Load-specific-artifacts-with-terms()$
        \STATE $Load-specific-artifacts-calculations()$
        \IF{$Artifact-detail$}
        \STATE Load-specific-artifact
        \STATE Visualization-of-artifact-with-similar-attributes
        \STATE Insights
\ENDIF
\ENDIF
\IF{Visualization}
        \STATE Load-Visualization()
        \STATE Load-Result-Categories()
        \STATE Insights
\ENDIF

\end{algorithmic}
\end{algorithm}

\begin{itemize}
    \item\textbf{Step1}: The web portal load the business-artifacts data, Feature Vectors and Machine Learning Algorithm Predictions.
    \item\textbf{Step2}: Choose the option to Analyze (Classification-based, Search-based and Visualization-based)
    \item \textbf{Step3a}: For Classification Based Choose the desired classification
    
    \item \textbf{Step3b}: For Search-based enter the search term for results
    \item \textbf{Step3c}: For Visualization Based 
    \item \textbf{Step4a}: Calculate the Evaluations for the classification
    \item \textbf{Step5a}: Load the Transactions with that classification filtered by correct and in-correct results segregation.
    \item \textbf{step6a}: Check transaction Details
    \item \textbf{step7a}: Visualizations of the artifact with different feature combination
    \item \textbf{Step4b}: Enter the search term
    \item \textbf{Step4c}: Calculate the Evaluations for the terms
    \item \textbf{Step5b}: Load Data containing the term and perform evaluations
    \item \textbf{Step6b}: Load the detail of specific artifact
    \item \textbf{Step7b}: Choose feature(s) to visualize  the classification of the artifact based with similar transaction set. 
    
\end{itemize}

The steps for the evidence discovery dashboard are described in the Algorithm \ref{alg:evidence}.

\subsection{Result Analysis and Feedback Loop}
To understand and analyze the system generated results,
data scientists need the support of non-complex systems, which are not another black-box to help understand the ML results. So the benefits of the proposed following sections in this approach are diversified and can support the business to improve the recognition of the machine learning system. 
\section{Feedback Loop}
The proposed reverse feature engineering component and the evidence-based dashboard act as the feedback loop to the system. With the provided support of these components, the machine learning algorithm performance can be improved. The improved version of the system with the enhanced confidence on the retrieved results help the organizations implement the machine learning algorithm with more confidence and will help improve the return over investment, an attractive point for the investors. 
The data scientist will better understand the designed algorithm and the behaviour prediction based on the historical data patterns. 
\subsection{Pipeline Intended Audience}
The intended audience of this proposed approach is data scientists, domain experts, business  administration, and other non-technical stakeholders of the business, as this front end is designed to support the non-technical users of the system with the appropriate visualization and details for the insights of the system. The supporting evidence is part of the system's already available dataset that helps provide a more detailed understanding. 


\section{Experiments and Evaluations}
\label{chap:motivating}

Money lending is among the significant and longest-standing financial institutes business products, and it facilitates people and organisations with an opportunity to attain their ambitions. Bank clients consist of numerous kinds of people with different preferences and motivations. During the 21st century, banking procedures are significantly improved due to the  advancement of technology. Recent developments have altered the banking landscape, and the potential of fraud and financial crimes has skyrocketed. Fintech, or Financial Technology, aims to supply and integrate the most cutting-edge technology in the financial industry. Sustainable Development Goals include strategies such as customising products depending on client preferences (SDGs). Addressing the customer needs leads to customer retention and ROI for all small and medium enterprises (SME). To discover more about their consumers, many businesses routed their resources to develop business operations. \cite{JenniferQ.Trelewicz2017BigSector}, \cite{Vives2017TheEconomy}. 
These recent changes in Fintech have attracted investments in intelligent system development to predict customer behaviour and predict their actions based on the existing data of the same or similar customers.  
AI-based systems are desirable, but the non-transparency of these systems has posed significant challenges. The existing limitation in the system has inspired us to conduct a scientific investigation to analyse the current issues of machine learning-based solutions in financial sectors and identify the venues to implement XAI approaches to address these issues.

\subsection{Fintech, Risk Analysis and Explanation}

Risk Analysis is a backbone of the financial institute as these studies and observations affect the evaluation of risk matrices and the policy design.  With the risk analysis of the loan aspirant, the organisation can find similar customers behaviour in the available data, predict future behaviour, find similar businesses, understand the company's current status and interpret the customer's contribution to the real economy. Lending Risk is related to the return over investment (ROI), a primary concern of the financier.  
To overcome this risk, the team of data scientists, risk analysts, and domain experts work together to analyse and improve the system's existing functionalities, identify the current approaches' shortcomings, and propose new solutions to combat the situations. 
\par
To analyse the  risk associated with customers, scorecard calculation is the critical component. Before the scorecard calculation, there are many preliminary steps. We have mentioned that previous loan history and transaction records can help understand the applicant's attitude towards handling loans and finances. To perform the risk analysis, the lending organisations use the score calculation mechanism. 
\par
To address the need of the hour and avoid the possible problems associated with the ML-based solutions. The organisations need to have explanation supported pipeline to answer the concerns and questions without adding any layer of ambiguity to the system. Studies are available with the Application of XAI in fintech, e.g., \cite{Bussmann2020ExplainableManagement} \cite{Bracke2019MachineAnalysis} and \cite{Fritz-Morgenthal2021FinancialAI} motivated us to conduct this study.

\subsection{Case Study: Bank Transaction Classification}
 The customer classification and risk grading based on the scorecard is a several staged complex process. In this section, we will walk through the example of one of the components of customer data, the Bank Transaction classification for decision making.
 
\subsubsection{Phase 1: Bank Transaction Extraction and Enrichment}
One of the significant and credible components of the loan applicant is the bank statement. It is an official bank generated document. Time-series data with short text, acronyms and numeric data involved. The applicant's data is large and tends to be skewed due to the enormity in the number of transactions performed monthly, making scrutiny and analysis on a large scale a problem of high computational complexity. Financial institutes request access to the bank transaction history for a period ranging from 1 to 12 months. Extracting information from a transaction description, location, and other components of a bank statement requires text mining techniques.

To pre-process the transaction data, two types of text mining algorithms are utilised.

\begin{itemize}
    \item At first stage, a low-level technique like cleaning, merging, or matching is applied.
    \item The application of high-level approaches would be the second method.
Lexical Feature tagging, Named Entities and Geo-locations are examples of techniques that may be used to extract keywords and phrases.
\end{itemize}  
Combining these methods would help to step forward to the feature vector generation for the ML model. 

\begin{figure}
    \centering
    \includegraphics[width=\textwidth]{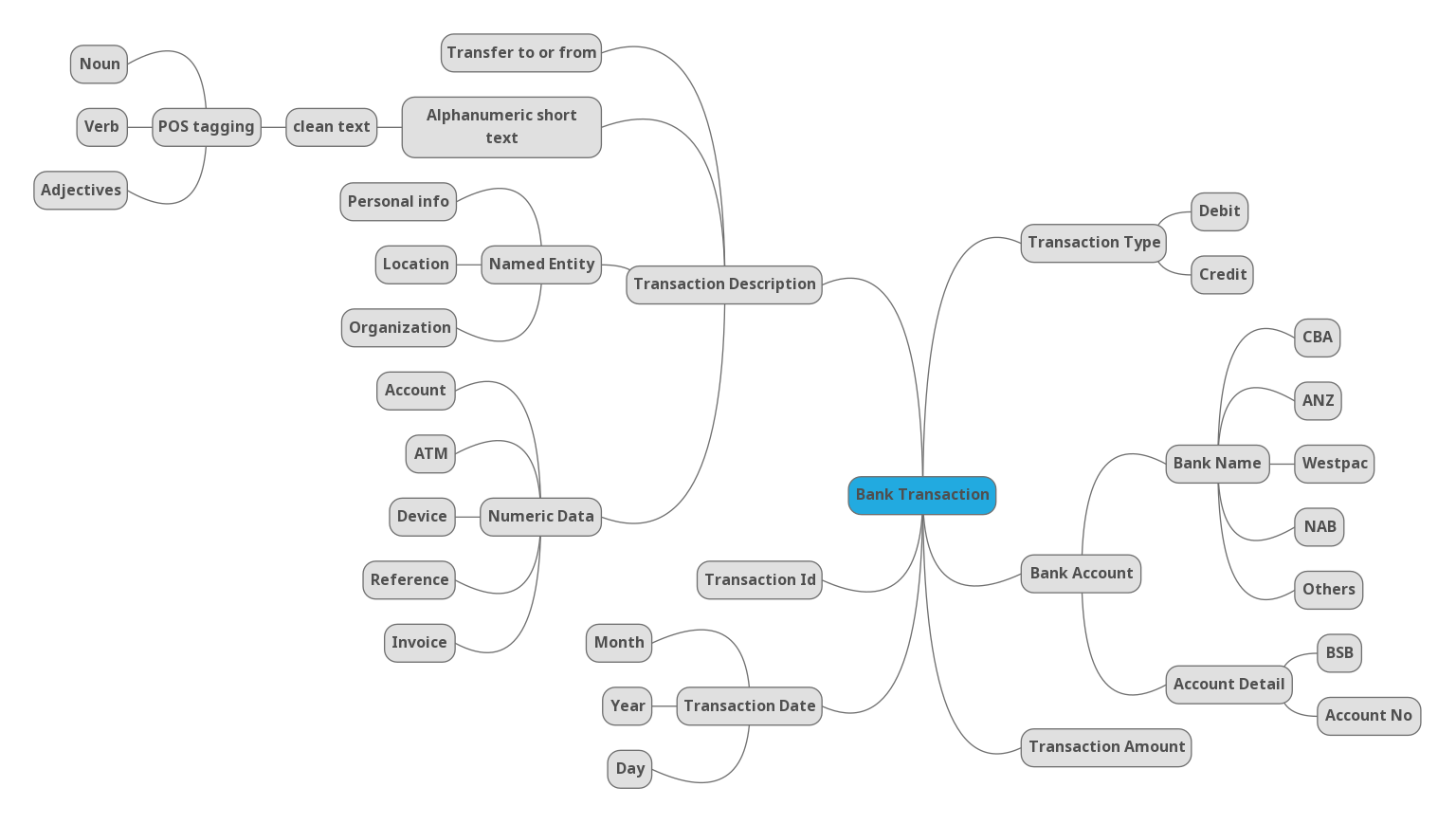}
    \caption{Components of the bank transaction and the possible options available for each component}
    \label{fig:transaction_component}
\end{figure}

The Figure \ref{fig:transaction_component} depicts the map of the transaction. This map summarises the various data fields available in the bank transaction. These data components lead us to understand which components of the transaction can enrich the system to predict the classification of the transactions better.

Although bank statements are complex and semi-structured, the innate capability to provide customer behaviour patterns over a certain period is also a significant source of truth about the customer's current financial situation and spending and earning pattern. The Time-series feature contributes to identifying the reoccurring payments that indicate the customers' commitments in utility bills, rents, groceries and income etc. Another important aspect of this legitimate document is availability over a while. It earns more scores to this component to understand the underlying pattern. The sample of the bank transaction is available in the Table \ref{tab:rawdata} to help the reader to understand the data. To have a robust system in place, the bank statements are enriched with domain knowledge from various sources to understand better. With the help of various Application Programming Interfaces (API)s the system is enhanced to relate the transactions. 

Few to name enrichment entities accessible are 
\begin{itemize}
    \item \href{www.abr.gov.au}{Australian Business Register }\footnote{www.abr.gov.au} provided by the Australian government
    \item \href{www.creditorwatch.com.au/}{Creditor Watch}\footnote{www.abr.gov.au} for credit history of business.
    \item \href{www.illion.com.au}{Illion}\footnote{www.illion.com.au} for credit history of individual business owners or sole traders. 
    \item \href{www.console.developers.google.com}{Google API} \footnote{www.console.developers.google.com} for location, contact details and reviews of the business.
     \item \href{www.immi.homeaffairs.gov.au}{VEVO} \footnote{www.immi.homeaffairs.gov.au} for citizenship status.
    \item \href{www.data.gov.au}{Australian Government data} \footnote{www.data.gov.au} for Gaming business registration number.
    \item \href{www.data.nsw.gov.au}{State Government data} \footnote{www.data.nsw.gov.au} for Industry registration number.
    \item Previous records of existing customers from the local database.
     \item \href{www.linkedin.com}{Linked In} \footnote{www.linkedin.com} for business and business owner's information.
\end{itemize}

\subsubsection{Phase 2: Machine Learning Model As a classifier}
Data engineers perform feature engineering after cleaning and transforming the data, and data scientists, with the help of tools, perform automatic feature selection for the model. Each selected feature is converted to a feature vector based on the data types, i.e., numerical, categorical, textual, visual, etc. These feature vectors are inputs of the ML system and classify each transaction. The transaction classification helps to calculate the Revenue and Expenses of the customer. It helps to calculate the profit of the loan applicant by subtracting the expense from the revenue. The machine learning algorithms are conventionally trained on historical time series data for prediction. 
The bank transactions are mainly of two types credit, and debit but further analysis of the transaction reveals they are of various types, including rent payment, bill payment, loans, fees, grocery, shopping, salary etc. Given the scenario, this is a multi-class classification problem. To solve the Multi-class Classification problem, various solutions exist in the literature. To fulfil our needs, we need a multi-classification system. Standard methods for multi-class classification mentioned in the literature are: 

\begin{itemize}
    \item K-Nearest Neighbors
\item Decision Trees
\item Naive Bayes
\item Random Forest
\item Gradient Boost
\item Neural Network 
\item Deep Neural Network
\end{itemize}

And the Multinomial probability function is required to solve the multi-class problem. As we have mentioned in the literature section, the system's innate ability to explainability reduces as we move towards the complex models. 

\textbf{Output Classes}

This study will focus on the Probabilistic Neural Network Model for the credit transaction classifications only.
The defined transaction classifications are 
\begin{itemize}
    \item Funding
    \item Income - Invoicing
    \item Income - Cash
    \item Income - Cheque
    \item Other 
\end{itemize} 
Other covers all the credit transactions except funding and income. e.g. internal bank transfers. 
So in total, we have five classes defined for the output. The system generates the probability of each type for each transaction, Sum  to 1, and the highest score is assigned as a final classification. 

\subsubsection {Phase 3: Feature Engineering}
The model takes a feature vector of a transaction to generate the probabilistic output of the system. Each feature vector is input into ML Algorithm to classify the transaction. The transaction data has 

\begin{itemize}
    \item Categorical input, e.g. Industry, Banks 
    \item Numerical Input, e.g. Transaction Amount
    \item Natural Language Text, e.g. Text description 
\end{itemize}

Various techniques are used to generate the feature vector from the data. For the categorical input like the industry category, the system has implemented one hot encoder. For the Numerical Input, we used scaling techniques for the dataset. 
For text processing, applied a group of techniques (e.g. text cleaning, text feature extraction, and text2vec conversion). The sample of the generated feature vector is shown below for better understanding.

\begin{lstlisting}[language=json,firstnumber=1,label={lst:input},caption=Input Data Example]
{   "ApplicationId": "32000",
    "IndustryCategory": "Meat",
    "BankAccounts": [
        {
            "Bank": "Suncorp Bank",
            "AccountName": "Cash Management Account",
            "AccountNumber": "123-456-789",
            "Transactions": [
                {
                    "Sha": "SHA_0001",
                    "Date": "2018-09-01T00:00:00",
                    "Amount": 7.47,
                    "Description": "DIRECT CREDIT A 123-56NSW",
                },
                {
                    "Sha": "SHA_0002",
                    "Date": "2018-09-02T00:00:00",
                    "Amount": 13.5,
                    "Description": "EFTPOS",
                },
                {
                    "Sha": "SHA_0004",
                    "Date": "2018-09-02T00:00:00",
                    "Amount": 15.5,
                    "Description": "EFTPOS",
                }
            ]
        },
        {
            "Bank": "Suncorp Bank",
            "AccountName": "Everyday Account",
            "AccountNickname": null,
            "AccountNumber": "123-456-788",
            "Transactions": [
                {
                    "Sha": "SHA_0003",
                    "Date": "2018-09-01T00:00:00",
                    "Amount": 1000.15,
                    "Description": "Commissions for xyz",
                }
            ]
        }
    ],
    }
\end{lstlisting}

The sample decision engine results for the bank transaction is shown below:

\begin{lstlisting}[language=json, firstnumber=1,label={lst:output},caption=Model Final Output Example]
{
    "Transactions": [
        {
            "Sha": "SHA_0001",
            "FinalClassification": "INCOME_CASH",
        },
        {
            "Sha": "SHA_0002",
            "FinalClassification": "INCOME_CASH",
        },
        {
            "Sha": "SHA_0003",
            "FinalClassification": "INCOME_CHEQUE",
        }
    ]
}
\end{lstlisting}

The final result listed above is based on the maximum probability predicted for the class. 
The complete model predicted values are also listed in the JSON listing below.

\begin{lstlisting}[language=json, firstnumber=1, label={lst:model probability},caption=ML Model Probabilities Example]]
{
    "Transactions": [
        {
            "Sha": "SHA_0001",
            "income_invoice": 0.5,
            "income_cash": 0.3,
            "funding": 0.1,
            "income-cheque": 0.1,
            "other": 0.0
        },
        {
            "Sha": "SHA_0002",
            "income_invoice": 0.4,
            "income_cash": 0.4,
            "funding": 0.0,
            "income-cheque": 0.1,
            "other": 0.1
        },
        {
            "Sha": "SHA_0003",
            "income_invoice": 0.3,
            "income_cash": 0.2,
            "funding": 0.0,
            "income-cheque": 0.5,
            "other": 0.0
        }
    ]
}
\end{lstlisting}

\subsection{Phase 4: Conventional Neural Network (CNN) Results Analysis}

The CNN generated results need further investigation by the domain expert, data scientist, and risk analyst to gain confidence in the predicted classification. In machine learning systems, accuracy is not enough to ensure the model performance for individual applications. Over time the previous data trends can be skewed, and the predictions are not correct. Without a thorough analysis of NN predictions, we cannot rely with confidence on the generated results. 

\subsubsection{Existing Limitation}

This stage highlights the limitation of machine learning and its acceptance in business due to the lack of answering the whys and hows of the system. 
Machine learning algorithms are considered non-trustworthy due to the inability to explain how the model retrieved a particular output. The algorithm cannot reveal the stages that out of many possibilities why the system opted for a specific route. The stochastic nature of the results also highlights that results for the same transaction can vary. This fact hinders the implementation of these algorithms alone on a large commercial scale. This is another challenge faced by the corporate sector that impedes the progress of the organisation. 

\subsubsection{Approach} 

For every limitation, we have opportunities. This limitation of the system motivated us to investigate and suggest a solution to address the limitations. To address the black-box model, one possible option is to unbox the NN. But to understand the data in a better way and to strengthen the Data Engineer and scientist to have faith in their technique, we opted for the explainability task. The XAI based approach will explain the insights and patterns of the dataset. This approach supports understanding the contribution of different components for the decision generation, which helps to optimise the input feature vector, and another significant contribution is providing the evidence. This evidence helps to understand where this instance is standing in the similar features of the dataset. These features of the approach help to support the business stakeholders without adding further any layer of the black box to understand the system. 

\subsubsection{Opportunity}

With the proposed approach, we are supporting data scientists and domain experts to understand the data with evidence and identify the patterns in the data. We are not limiting the explanation to a specific set. The evidence matching is available for the available dataset, and domain experts can choose the desired window for understanding. The interactive nature of the evidence tuneable dashboard facilitates understanding prediction. 

\subsection{Explanation of Results to Business Stakeholders}

Revisiting the classified transactions are still an overhead for the bookkeepers to correct the decisions made by the system and then report back so the intelligent algorithm team ungraded the system performance by improving the training set and reinforcing learning. But this process is not concrete and is not enough feedback to data scientists for enhancing the system contribution to improve the system. This section highlights the existing gap and the need for a one-stop dashboard to analyse the results and data generated by the system. 
To improve the system and reduce the load on the bookkeeper, we can provide an explanation dashboard for the domain experts( Risk Analyst/ Machine Learning engineer) to compare the transaction with similar transactions and analyse the generated results of the system critically. This feedback will help the domain expert visualise the data and the similarity of the current entity with the existing ones in real-time. 

\subsubsection{Use Case 1: Inability of how a particular classification was assigned?}

Based on the bank transaction classification, certain significant amount credit transactions are not classified as income which reduces the revenue of the specific loan applicant. 
Now the customer wants to challenge the loan rejection decision provided. The domain expert and risk analyst wish to review the decision, and to do that, they need to understand how the model  decided to avoid such problems for the next time.  But the model is unable to answer this inevitable question. The machine can only help to pass the feature vector of the applicant again and get the classification. And for being a stochastic system, the classification result for this iteration can be varying. 
The data analyst and multiple stakeholders need to visualise and analyse the algorithm's results to develop confidence in the classification. They need to ensure that data is not biased, and XAI can help to reduce bias \cite{Nobel2021WhyLearning}.
This scenario motivates us to conduct a critical analysis of the results. Currently, the analyst cannot understand the generated classification of the system as the ML algorithm is a black box. The reasons for the output is not predictable and has no trace to move back. We need a mechanism to ensure the data is not biased and we have a reasonable amount of data for all classes to avoid skew.

\subsubsection{Use Case 2: Inability to answer Why this classification?} 

The machine learning models are black boxes, and they are complex too. Unboxing the logic is not straightforward. These models cause another hindrance for the domain expert in terms of explaining to the other stakeholder. The domain expert cannot answer why the machine assigned a particular classification to a specific transaction. Apart from this, similar transactions are available in training data. 
A convincing argument to satisfy the stakeholder and raise the company's trust in the particular algorithm is essential. 
This scenario again highlights the lacking of the existing system. 
With the machine learning algorithms, a question is very trivial, How does the machine generate these results? 
For the data scientist analysing a result is always critical. In real life, machine learning models cannot solely rely on accuracy and sensitivity scores. We need to analyse and justify the decisions made for each transaction, and for that, we need a mechanism to dig deeper into the system. The lacking of this attribute also questions the fairness of the decision generated.  

\subsubsection{Use Case 3: Inability to help understand and compare results of two same classification}.
For example, a loan aspirant has \$2000 profit as per the system decision. We can offer him a loan of \$50,000, whereas another user has a net profit of \$3000 per month,  but the system rejects the loan application. There can be varying reasons behind this problem. One of the reasons can be the identified transactions have a high-risk transaction, e.g. money came from some non recommended resources or money spent on illegal activities. But ML-based systems are unable to provide answers to such questions. The complexity adds hindrance to the transparency. 
ML model will be silent to answer the user's concern about why two similar transactions got two different classifications? Which particular transaction feature contributed to extracting the feature. As well as the historical patterns played a significant role to retrieve the decision from the model.  In such a scenario, the companies can lose potential customers, and it raises a question mark on the fairness of proposed answers of the ML system. 

\begin{figure}
    \centering
    \includegraphics[width=\textwidth]{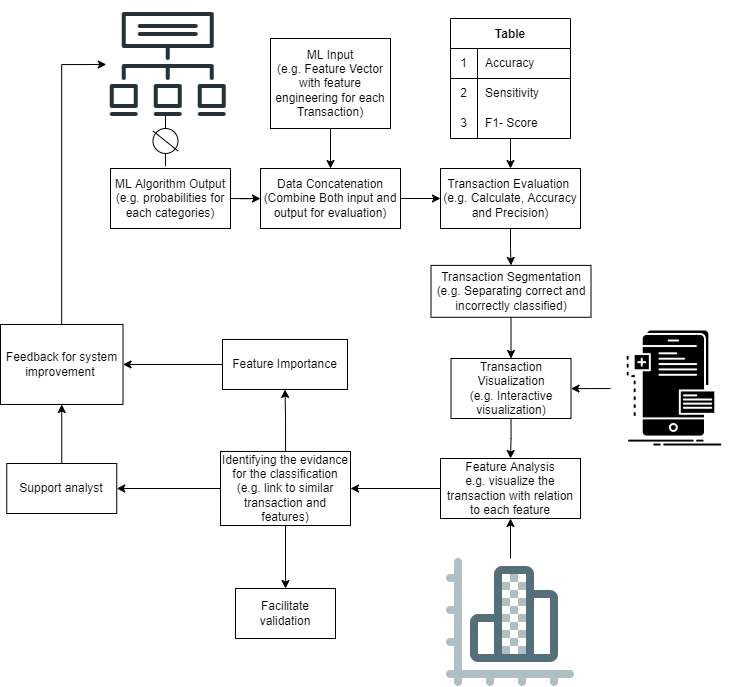}
    \caption{This is an instance of the evidence-based system execution using the bank transaction classification data. This helps to understand the process followed to generate I/O of the system and how interactive dashboard supports human.}
    \label{fig:stage_transaction}
\end{figure}

These all questioned motivated us to conduct this study.
The overall visualisation of the proposed motivating scenario is presented in Figure \ref{fig:stage_transaction}


\subsection{System Setup}

This experiment is mainly conducted on the \href{https://databricks.com/}{Azure Databricks Platform} using the parallelization approach to reduce the computational time as the filtered raw data at hand was more than 8GB. The ML output and the input feature vectors are then sent to the Evidence-based dashboard for visualization purposes, it was developed using the Streamlit package in python.
The web app interacts with the data retrieved from the Azure Databricks and updates interactively. The results are fed back to the Azure Data.

\subsubsection{Feature Vector} 
The feature vector set developed and used for the machine learning is shown in the Table \ref{tab:featurevector} and these features are input to the ML part and then are used for the interactive dashboard for evidence-based analysis of the results. The Text2vec component is not shown to make this data snippet more meaningful to the reader. For the machine understanding, the text was converted to a vector.
\subsection{Machine Learning Algorithm Results}
The machine learning algorithms results for each transaction classification are used to work with the feature vector to develop the evidence-based systems. These results are used to calculate the accuracy, sensitivity, and F1 scores for the classification.  
\subsubsection{Analysis}
The feature vector, machine learning results, and calculations are combined to generate the evidence-based system for visualization and reverse feature engineering based on feature importance.


This section presents the experimental setup for the system and discusses the application design with the demonstration. 

\subsection{Raw Dataset}
This study used data from \href{www.prospa.com}{Prospa Group Ltd}, Sydney, Australia based private financial institution. Individual small enterprises that applied for a small business loan in the last six months are included in the dataset. The total number of transactions are  19694039 for bank transactions of 14828 customers. All client information is replaced with digits in the range of 1 to 14828 to ensure security and privacy. A range of information is gathered on customer's authorization to access their bank statement. 
However, we need limited information from the data source to explain and reflect on the system design for this study. 
Our data set has: 

\begin{itemize}
    \item Customer Id: An integer assigned for each customer. 
    \item Bank name: Financial institution of account holder.
    \item Transaction date: The calendar date of the transaction in DateTime format.
    \item Transaction type: In our case, we are only using credit transactions. 
\item Transaction description: A brief description of the transaction's objective.
\item Transaction amount: The amount of money involved in each transaction in float format.
\item Industry class: Business industry of the applicant.
\end{itemize}
Table \ref{tab:rawdata} represents a sample of bank transactions from the raw dataset. 
Notably, we are using the applicants from Australia in this study, and the amount presented is in Australian dollars.

\begin{table*}[htbp]\scriptsize
    \centering
    \caption{Sample of Raw dataset for understanding}
    \begin{tabular}{|c|c|c|c|c|c|}
    \toprule
         \textbf{\makecell{Customer \\ Id}} & \textbf{\makecell{Bank \\ Name} }& \textbf{\makecell{Transaction \\ Amount}} & \textbf{\makecell{Transaction \\ Date}} & \textbf{\makecell{Transaction \\ Description} }& \textbf{\makecell{Industry \\ Class} }\\
         \midrule
         1& ANZ & 280.97 & 2019-06-09 & \makecell{EFTPOS \\TRANSACTION } & Hospitality  \\
         \midrule
         2  & NAB & 802.47 & 2020-03-04 &\makecell{INTER-BANK \\ CREDIT Wages \\Bank}	& \makecell{Building and Trade}	\\
         \midrule 
         3 & NAB & 150 & 2020-11-02	& \makecell{TRANSFER \\CREDIT ONLINE \\Linked transaction \\ CONSTRUC} & Hospitality\\
         \midrule
         4 & NAB & 626 & 2019-09-18 & \makecell{McDonald \\ 3xxxcx6} & \makecell{Professional \\services} \\
         \bottomrule
         
    \end{tabular}
    
    \label{tab:rawdata}
\end{table*}

\begin{table*}[htbp]\scriptsize
    \centering
    \begin{adjustbox}{angle=90}
    \begin{tabular}{|c|c|c|c|c|c|c|c|}
    \toprule
         \textbf{ \makecell{Cus \\ Id}} & \textbf{Bank}& \textbf{\makecell{Tran. \\ Amount}} & \textbf{\makecell{Year}} & 
         \textbf{\makecell{Month}}& \textbf{\makecell{Date}} & \textbf{\makecell{Transaction. \\ Description} }& \textbf{\makecell{Industry \\} }\\
         \midrule
         22931 & 2 & 4.453 & 2020 & 06& 09 & \makecell{['direct', 'credit', '2xxxx3', 'myob', 'pay','by', '000003']} & 1  \\
         \midrule
         22241 & 33 & 2.80 & 2020 & 06 & 09 & \makecell{['miscellaneous', 'credit', 'internet', 'payment']} & 1 \\
         \midrule
         27909& 46 & 0.11 & 2021 & 01 & 04 &\makecell{['deposit', 'payment','bank']} & 4 \\
         \midrule
         22819 & 25&8.8 & 2021 & 02  & 15& ['payment', 'from', 'abc', 'xyz'] & 4\\
         \midrule
         28707 & 19 & 0.5 &2021 & 04& 20& ['season', '2xxx45', 'bank', '0xxxx'] & 7\\
         \midrule
         27414 & 3 & 1.4 & 2021 & 01 & 14 & ['name', 'else', 'person', 'transfer'] & 2  \\
         \midrule1
         25459 & 51 & 0.4 & 2020 & 09 & 29 & ['payment', 'received', 'mangoes', 'transfer', 'amount','for','bbbb'] & 4 \\
         \midrule
         21474 & 117 &1.5 & 2021 & 08 &20& ['miscellaneous', 'credit', 'internet', 'payment', 'purchase'] & 6\\
         \bottomrule
    \end{tabular}
    \end{adjustbox}
    \caption{Sample of features vector prepared on collected data after cleaning}
    \label{tab:featurevector}
    
\end{table*}


\subsection{Evaluation}
Standard matrices are used to analyse the system's intended pipeline. These matrices are tailored to ensure that the system is implemented effectively. 
The system is a multi-class classification task for machine learning component assessment. We utilized accuracy \ref{eq:1}, Precision \ref{eq:2}, Cohen-Kappa \ref{eq:3}, Recall \ref{eq:4}, F-measures \ref{eq:5} and Support. The number of successfully categorised transactions by the PNN divided by the total number of predictions is \textbf{Accuracy} and ratio of correctly classified transactions to total transactions is \textbf{Precision}. \textbf{Cohen-Kappa} gives a measure of a proposed system categorisation and actual category agreement to a mutually exclusive class. \textbf{Recall} is the ratio of correctly categorised transactions to relevant transactions, and \textbf{F-measure} is the harmonic mean of Precision and Recall. \textbf{Support} is the number of actual occurrences of the class in the test data set. The following formulas show the mathematical expressions for these measures.

\begin{equation} \label{eq:1}
    Accuracy = \frac{(TP + TN)}{(TP + TN + FP + FN)}
\end{equation}
\begin{equation} \label{eq:2} 
Precision = \frac{TP} {(TP + FP)}
\end{equation} 
\begin{equation} \label{eq:3}
Cohen-Kappa = \frac{(p - q)} {(1 - q)}
\end{equation}
\begin{equation} \label{eq:4}
    Recall =\frac{TP}{(T P + F N)}
\end{equation}
\begin{equation} \label{eq:5}
    F-measure =\frac {(2 * Precision * Recall)}{(Precision + Recall)}
\end{equation} 
 
Here, TP, TN, FP, FN stands for True-Positive, True-Negative, False-Positives, and False-Negatives, respectively. The number of mutually categorised transactions divided by the total number of transactions is p, and the number of accurate categorisation minus the wrong classification is q.
With the classic ML-based approach, achieved results are mentioned in Table \ref{tab:result}. The overall accuracy of the system is 96.5\%.

\begin{table}
\centering
\caption{The results of Machine Learning Algorithm in terms of accuracy, precision, recall, F-measure and sensitivity}
\begin{tabular}{|c|c|c|c|c|c|}
\toprule
    Classification  &  Accuracy & Precision &  Recall & F-measure & Sensitivity \\
    \toprule
     Funding &
     
    96.5\% & 90.79\%& 	80.11\% & 19.5\% & 67.25\%

\\
     Income - Cash & 95.9\% &97.59\% &93.17\%	&96.93\% &
     65.72\%
\\
     Income - Invoice & 96.5\% &97.88\% 	& 34.46\% & 20.83\% & 20.83\%
\\ 
     Income - cheque & 95.4\% &97.95\% & 92.95\% &	96.93\% &94.28\%
\\
    Other & 96.5\% & 95.04\%& 92.95\%&99.12\%	&97.04\%\\

\bottomrule
\end{tabular}

\label{tab:result}
\end{table}

\subsubsection{Scalability Study}

This system is scalable and can extend beyond credit transactions, and is available to a live system to give results on the live system. We can configure the data bricks server for the live visualization of the results as well. This will improve the execution time as well as the transparency of the results with interactivity. 
The live data handling will reduce the cost as currently, we are caching the data for the visualization purpose, which will decrease with the live cluster data handling and delta tables. With this cluster-based support, the result loading time will reduce to 2 milliseconds, from 1 minute. However it increases with the increase in data size.


\section{Conclusion and Future Directions}
\label{chap:discussion}

\subsection{Concluding Remarks}

It is critical to guarantee that financial services are up to date and simple to use in order to stay up with technological advancements. Understanding and analysing risk is one of the most crucial parts of the financial sector. Commercial enterprises, such as banks, may have a comprehensive understanding of their clients and products by using an effective and efficient risk metric. This industry may flourish with improved customer satisfaction by adopting insights technology to support AI decision's. Customers are classified into the appropriate risk category by recognizing informative criteria e.g., risk score, client an d transaction type, revenue and expense predictions. Financial firms can decide the kind of services and products that will benefit both the consumer and the bank based on risk classification. With the added component of insights-with-evidence will support the financial sector to improve the performance and growth of the organization and to improve the acceptability of the black box models of with system without enhancing the black box capability of the system. 
The paper help understand the machine learning model with the evidence based approach through interactive dashboard and reverse feature engineering~\cite{irecruit}. The proposed evidence based pipeline is a set of processes which are way forward to support the organization with the tool to ease the daily interaction with the machine learning system results and aid the results with powerful visualization system.

\subsection{Future Work}

We have developed a working prototype and tested it on financial data. However, in the light of detailed survey and tools presented, the prototype can be further developed. Below are some possible direction to expand the work presented in this paper. 

\subsubsection{Explanation Techniques Implementation}
One of the potential next steps for enhancing the project is to have multiple famous explanation techniques \cite{Kuhn1953ContributionsAM-28}, \cite{Ribeiro2016WhyClassifier} implemented in the system. 

The system can provide the techniques to choose the necessary process from the dashboard to compare the results of different features in the same setup. It will help to grow the user's confidence in the proposed method, as well as the same platform will provide options for added explanation algorithms. 
This component addition will make it an all in one solution for the explanation generation for the bank transaction classification system. 
With the implementation of these techniques, the evaluation will be a vital step to understand the stance of the explanation. 
As this is the case of individual transaction classification, we can use local transaction techniques first and then move towards global explanations. 

\subsubsection{Multiple Machine Learning Algorithm Results Comparison}

The machine learning algorithms need to upgrade over time, and the scenario we have considered is time-series data. There can be possible reasons for upgrading the model training data from data drift to inevitable policies and laws. Either we are manually upgrading the ML algorithm or using a continual approach. We need to compare the results of the processes. We  can extend this dashboard to visualize the effects of different pipeline versions for better understanding and reaching the decision. We need an efficient tech and human stack to improve better system training and compare the results. This dashboard can be an added support for the team and tools for the model visualization and improvement journey.
We also need to evaluate the confidence score for each transaction result with the machine learning model comparison. We need to develop a mechanism that can also calculate how much confidence score was assigned to the transaction classification at each instance by the machine learning algorithm. It will help to trust the decision based on a certain confidence threshold; otherwise, we need to understand why the low score. 

\subsubsection{Various Segregation in Data}
In this case study, we have just used incorrect and correct segregation. We can extend it beyond the accurate and inaccurate levels to filter based on a certain threshold for each transaction. We can also identify the close match transactions based on the verge for the improved understanding and readability of the transactions. With various types of segregation, we will be able to identify the overlapped or closely matched patch with the available dataset. Furthermore, these segregation techniques will help understand the nearest neighbours and their effects on the final transaction.

\section*{Acknowledgements}
- I acknowledge the AI-enabled Processes (AIP\footnote{https://aip-research-center.github.io/}) Research Centre and Prospa\footnote{https://www.prospa.com/} for funding My Master by Research project.

\bibliographystyle{abbrv}
\bibliography{ms}

\end{document}